\documentclass[journal]{IEEEtran}
\IEEEoverridecommandlockouts
\usepackage{amsfonts}
\usepackage{algorithmic}
\usepackage{textcomp}
\usepackage{newfloat}
\usepackage{listings}
\usepackage{lipsum}
\newcommand{\cmark}{\ding{51}}
\usepackage{biblatex}
\usepackage{enumerate}
\usepackage{pifont}
\usepackage{times}
\usepackage{epsfig}
\usepackage{graphicx}
\usepackage{amsmath}
\usepackage{amssymb}
\newcommand{\xmark}{\ding{55}}
\newcommand{\ie}{\textit{i.e.}}

\usepackage{subcaption}
\usepackage{indentfirst}
\usepackage{relsize}
\usepackage{siunitx}
\usepackage{array,multirow,adjustbox}
\makeatother
\usepackage{multirow}
\usepackage{tikz}
\usepackage{amsthm}
\usepackage{makecell}
\usepackage{booktabs}
\usepackage{color}
\usepackage{colortbl}
\usepackage{xcolor}
\usepackage{array}
\usepackage{listings}
\usepackage{pgfplots}
\usepackage{fancyhdr}

\usetikzlibrary{pgfplots.polar}
\pgfplotsset{compat=1.18}
\usepackage[linkcolor=blue, citecolor=blue,  colorlinks=true]{hyperref}
\definecolor{dkgreen}{rgb}{0,0.6,0}
\definecolor{gray}{rgb}{0.5,0.5,0.5}
\definecolor{mauve}{rgb}{0.58,0,0.82}

\def\BibTeX{{\rm B\kern-.05em{\sc i\kern-.025em b}\kern-.08em
    T\kern-.1667em\lower.7ex\hbox{E}\kern-.125emX}}
\makeatletter

\newcommand{\linebreakand}{%
  \end{@IEEEauthorhalign}
  \hfill\mbox{}\par
  \mbox{}\hfill\begin{@IEEEauthorhalign}
}

\makeatother

\begin{document}
\title{Learning Motion and Temporal Cues for Unsupervised Video Object Segmentation}
\author{Yunzhi Zhuge,~\IEEEmembership{Member,~IEEE}
        Hongyu Gu,
        Lu Zhang,
        Jinqing Qi,~\IEEEmembership{Member,~IEEE}
        and Huchuan Lu,~\IEEEmembership{Fellow,~IEEE}


\thanks{Manuscript received 2 January 2024; revised 23 April 2024; accepted
20 June 2024. \textit{(Corresponding author: Lu Zhang.)}}
\thanks{Y. Zhuge, H. Gu, L. Zhang, J. Qi and H. Lu are affiliated with the School of Information and Communication Engineering, Dalian University of Technology, China (e-mail: zgyz@dlut.edu.cn; guhongyu@mail.dlut.edu.cn; zhangluu@dlut.edu.cn; jinqing@dlut.edu.cn; lhchuan@dlut.edu.cn).}}


\maketitle

\begin{abstract}
In this paper, we address the challenges in unsupervised video object segmentation (UVOS) by proposing an efficient algorithm, termed MTNet, which concurrently exploits motion and temporal cues. Unlike previous methods that focus solely on integrating appearance with motion or on modeling temporal relations, our method combines both aspects by integrating them within a unified framework. MTNet is devised by effectively merging appearance and motion features during the feature extraction process within encoders, promoting a more complementary representation. To capture the intricate long-range contextual dynamics and information embedded within videos, a temporal transformer module is introduced, facilitating efficacious inter-frame interactions throughout a video clip. Furthermore, we employ a cascade of decoders all feature levels across all feature levels to optimally exploit the derived features, aiming to generate increasingly precise segmentation masks. As a result, MTNet provides a strong and compact framework that explores both temporal and cross-modality knowledge to robustly localize and track the primary object accurately in various challenging scenarios efficiently. Extensive experiments across diverse benchmarks conclusively show that our method not only attains state-of-the-art performance in unsupervised video object segmentation but also delivers competitive results in video salient object detection. 
These findings highlight the method's robust versatility and its adeptness in adapting to a range of segmentation tasks. Source code is available on  \href{https://github.com/hy0523/MTNet}{https://github.com/hy0523/MTNet}.
\end{abstract}

\begin{IEEEkeywords}
Unsupervised video object segmentation, video salient object detection, temporal transformer, multimodal fusion.

\end{IEEEkeywords}

\section{Introduction}
Video Object Segmentation (VOS) is a fundamental task in computer vision, which involves precisely locating and segmenting objects in all frames of a video. Different from semi-supervised video object segmentation (SVOS) which relies on ground-truth masks in the first frame to perform tracking and segmentation in subsequent frames, unsupervised video object segmentation (UVOS) aims to adaptively segment objects without any human intervention. More specifically, the intrinsic value of UVOS lies in its autonomous segmentation capability, crucial for real-time applications such as autonomous driving~\cite{hu2022goal} and video editing~\cite{zeng2020learning}. In these contexts, rapid and accurate analysis of video content is essential, particularly as it does not rely on manual annotations.
Nevertheless, achieving accurate segmentation in complex scenarios continues to pose a substantial challenge in the field of UVOS.

\begin{figure}[t]
\centering
\vspace{-2mm}
\begin{tabular}{@{}c}
\includegraphics[width=0.98\linewidth]{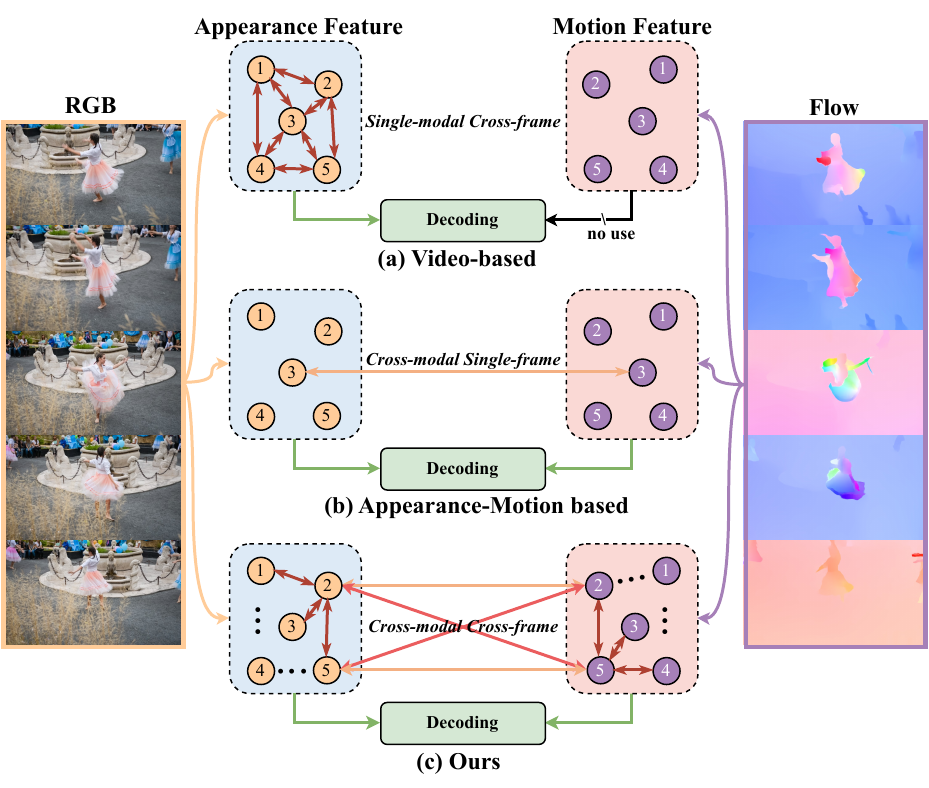} \\
\end{tabular}
\vspace{-3mm}
\caption{Three distinct methodologies for unsupervised video object segmentation (UVOS). (a) The video-based approach consolidates temporal information to enhance feature representation across frames.  (b) The appearance-motion-based method utilizes optical flow for motion guidance and compensation. (c) Our proposed method innovatively synthesizes both temporal and motion information within a cohesive framework, facilitating the transfer of cross-modal and cross-frame knowledge.}
\vspace{-4mm}
\label{figure:motivation}
\end{figure} 

In the domain of unsupervised video object segmentation, several methods naturally associate the near frames via dynamic attention mechanism~\cite{wang2019learning}, graph neural networks~\cite{wang2019zero} and pyramid constrained self-attention~\cite{gu2020pyramid}.  However, these video-based approaches may still struggle to precisely locate the main object, primarily due to the absence of motion information. This type of information is crucial for imparting prior knowledge that helps differentiate object significance and determine trajectories in unsupervised video object segmentation (UVOS). As a result, high-interference objects can significantly impair the performance of existing algorithms when dealing with complex videos. Moreover, while some methods incorporate frame-to-frame relationships to some extent, their effectiveness remains limited in handling videos of extended duration.

Currently, motion-appearance modeling forms the foundation for most advanced unsupervised video object segmentation (UVOS) methods. These methods employ optical flow to guide motion tracking, while extracting appearance cues from original images. By combining these two modalities, the model can capture the characteristics of the primary object, as well as its movements throughout the video.
Although this integrative approach has been proven to be effective in UVOS across several datasets, there remains considerable scope for improvement. Firstly, those fusion mechanisms typically depend on sophisticated operations to reach superior performance, leading to an increase in model parameters that challenges the feasibility of deployment on practical devices. Additionally, some methods~\cite{zhou2020motion,zhen2020learning,ji2021full,yang2021learning} incorporate a post-processing step using Conditional Random Fields (CRF)~\cite{krahenbuhl2011efficient} to refine object boundaries, further increasing the computational burden of already resource-intensive algorithms. Secondly, they do not explicitly model the video information, which is crucial for tracking the primary object in scenarios characterized by rapid object displacement and frequent occlusions.

After observing the limitations of the methods mentioned above, it is natural to make the suspicion: \textit{what makes for an UVOS algorithm capable of handling various challenging scenarios efficiently?} 
Generally, a state-of-the-art tracker ideally exhibits the following attributes:
\begin{enumerate}[1)]
\item Accurate localization of the primary object within a video sequence,
\vspace{1mm}
\item Robustness against challenges presented by complex video sequences, particularly those involving frequent object occlusions and disappearances.
\vspace{1mm}
\item Efficiency and adaptability to diverse hardware constraints, ensuring rapid execution across various devices.
\end{enumerate}



Motivated by the aforementioned principles, we present a methodical and efficient algorithm, referred to as MTNet, which simultaneously leverages motion and temporal cues to address the complex unsupervised video object segmentation task. 
As is shown in Fig.~\ref{figure:motivation}, different from previous method that primarily focus on integrating appearance with motion or modeling temporal relations independently, MTNet combines these aspects within a cohesive framework.  
More specifically, the Bi-modal Fusion Module is meticulously designed to seamlessly integrate appearance and motion features during the encoding phase, thereby creating a more comprehensive representation. This fusion is pivotal for the UVOS task, as it significantly enhances the model's ability to interpret complex visual narratives without manual labeling,  establishing a fundamental component of autonomous systems.
To fully capture the intricate long-range contextual dynamics and embedded information within videos, we introduce the Mixed Temporal Transformer, which enhances  inter-frame interactions effectively throughout a video clip. Additionally, the Cascaded Transformer Decoders work in concert across various feature levels, meticulously refining segmentation masks. This progressive refinement is crucial for achieving the granularity required in high-stakes scenarios, such as real-time surveillance and on-the-fly video content generation.
Collectively, these innovations position MTNet not merely as an end-to-end solution but as a robust, sophisticated framework that leverages temporal and cross-modality insights. It excels in localizing and tracking primary objects with high fidelity, cementing its significance in advancing the state-of-the-art in UVOS and its practical deployment in industry-relevant applications.

We conduct experiments on a wide range of unsupervised video object segmentation (UVOS) datasets and video salient object detection (VSOD) datasets, making several noteworthy observations:
\begin{enumerate}[1)]
    \item MTNet exhibits stronger capability in handling long-term and motion occlusions, solidifying its position as a robust UVOS solution relevant for both academic research and real-world applications.
    \vspace{1mm}
    \item Across a diverse set of UVOS and VSOD benchmarks, MTNet consistently achieves state-of-the-art performance, underscoring its versatility and effectiveness across multiple tasks and scenarios.
        \vspace{1mm}
    \item A notable strength of MTNet lies in its inference speed, which is attributable to its efficient architectural design. Specifically,  MTNet operates at 43.4 frames per second (fps) on a 2080Ti GPU, emphasizing its capability for real-time implementations and its suitability for environments where computational resources are constrained.
\end{enumerate}


\section{Related Work}
\subsection{Unsupervised Video Object Segmentation}
Unsupervised Video Object Segmentation (UVOS)~\cite{TNNLS_MASK_RK, TNNLS_zhou2021self, TNNLS_yin2021directional, TNNLS_li2022self, zhou2022survey, lu2020learning} represents a specialized subset of video object segmentation that operates independently of human interaction during inference. The primary challenge in UVOS stems from the need to identify dominant objects within complex and visually distracting environments without any prior annotations or explicit knowledge.
In the realm of deep learning, particularly following the fully convolutional network (FCN)~\cite{long2015fully}, which revolutionized semantic segmentation with per-pixel predictions, UVOS has experienced substantial advancements and progress. 
Drawing inspiration from non-local networks~\cite{wang2018non}, several approaches, including COSNet~\cite{lu2019see}, AGNN~\cite{wang2019zero} and F2Net~\cite{liu2021f2net}, have been developed to model inter-frame correspondence and achieved a more comprehensive understanding of video content.
On the other hand, optical flow supplies vital motion information that aids in the localization and differentiation of primary objects, thereby enriching the UVOS process with crucial dynamic cues. 
MuG~\cite{lu2020learning} develops a multi-granularity VOS framework that leverages prior knowledge and multiple levels of granularity from frame-specific to whole-video to enhance video object segmentation.
MATNet~\cite{zhou2020motion} presents a two-stream interleaved encoder, providing a motion-to-appearance pathway for information propagation and a Motion-Attentive Transition Module for feature selection.  
Employing a full-duplex strategy, FSNet~\cite{ji2021full} designs a relational cross-attention module and a bidirectional purification module to effectively fuse appearance and motion information.
HFAN~\cite{pei2022hierarchical} introduces a hierarchical feature alignment network that aligns appearance-motion features with primary objects and adaptively fuses them to enhance performance. Although these two-stream methods demonstrate satisfactory performance in certain scenarios, they struggle to track primary objects without temporal contexts, particularly in intricate occlusion scenes.

\subsection{Video Salient Object Detection}

In the realm of computer vision, Video Salient Object Detection~\cite{wang2017saliency, wang2017video, fan2019shifting,wang2021salient} closely parallels to Unsupervised Video Object Segmentation, which aims to autonomously identify and delineate regions of significance within video sequences. Earlier approaches rely primarily on hand-crafted features and heuristic-based methods for discerning salient objects. For example, Wang et al.~\cite{wang2017saliency} presents a geodesic distance-based approach for video saliency, enhancing unsupervised video object segmentation with robust, temporally consistent superpixel labeling. This method exemplifies early efforts to identify salient objects in video sequences.

The introduction of deep learning models has shifted the paradigm substantially, resulting in significant performance enhancements. 
Wang et al.~\cite{wang2017video} introduces FCN for video saliency detection, tackling data scarcity and enhancing training speeds. By integrating static and dynamic saliency networks, this approach set new performance benchmarks. Meanwhile, SSAV~\cite{fan2019shifting} contributes the Densely Annotated VSOD (DAVSOD) dataset, comprising 23,938 frames, with precise ground-truths derived from human eye-fixations. Recognizing the saliency shift challenge, SSAV conducted an extensive evaluation of 17 VSOD models across various datasets, amounting to 84K frames, providing a thorough review of their performance. Additionally, CSANet~\cite{TNNLS_ji2020casnet} introduces a cross-attention encoder-decoder in the Siamese framework. This model enhances both intra-frame accuracy and inter-frame consistency by leveraging self-attention and cross-attention mechanisms. CoSTFormer~\cite{TNNLS_liu2023learning} emphasizes the overlooked complementarity in spatial-temporal (ST) knowledge of earlier works. It innovatively decomposed the ST context into short-global and long-local segments, leveraging transformers for assimilating complementary ST contexts.

\begin{figure*}[t]
\centering
\begin{tabular}{@{}c}
\includegraphics[width=0.98\linewidth]{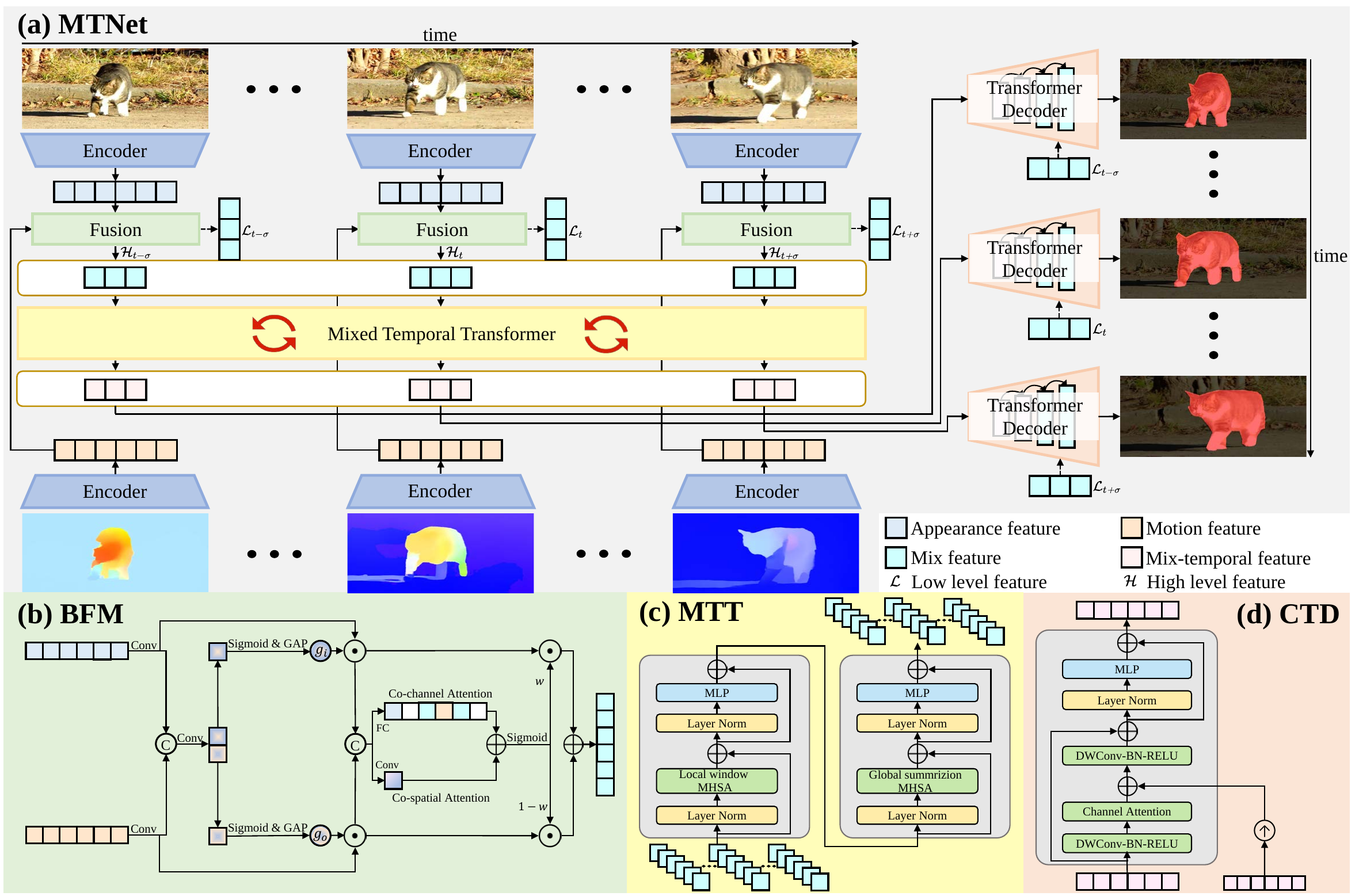} \\
\end{tabular}
\vspace{-1mm}
\caption{(a) The proposed MTNet pipeline utilizes $t$ frames of images and flow maps as input to extract multi-level features. These features at each level are fused by the (b) Bi-modal Feature Fusion Module. Subsequently, the temporal modeling of high-level features are achieved through the (c) Mixed Temporal Transformer. Finally, the output masks are generated using the (d) Cascaded Transformer Decoder.}
\label{figure:pipeline}
\vspace{-2mm}
\end{figure*}

\subsection{Feature Fusion in Segmentation} 
Feature fusion, which involves integrating multi-source data, has gained substantial interest in the research community recently. The complementary relationship among diverse inputs can significantly enhance performance in various computer vision tasks, with a notable impact in segmentation~\cite{TNNLS_zhao2023mitigating,liu2022cmx,wang2022multimodal,deng2021feanet}.  In particular, researchers have explored a wide range of modalities, such as depth~\cite{liu2022cmx,wang2022multimodal} and optical flow~\cite{rashed2019optical}, to complement traditional RGB information. Methodologically, these approaches can be generally divided into two main categories: convolution-based techniques and transformer-based strategies. 
In~\cite{zhuang2021perception}, the authors effectively integrate appearance information from RGB images with spatio-depth details from point clouds using residual-based fusion modules. Capitalizing on the proficiency of vision transformers in modeling long-range dependencies and global contexts, numerous recent studies have also employed self-attention or cross-attention mechanisms to fuse features from multimodal inputs. 
CMX~\cite{liu2022cmx} introduces a cross-attention-based framework with a versatile design that has been effectively generalized across five multimodal semantic segmentation tasks. TokenFusion~\cite{wang2022multimodal} employs a two-step process that initially prunes uninformative tokens within each modality, followed by the substitution of these tokens with aligned features from other modalities.
Inspired by these approaches, we propose Bi-modal Fusion Module to effectively integrate appearance and motion features. To the best of our knowledge, this constitutes an early attempt to apply such an operation in the context of unsupervised video object segmentation.

\subsection{Video Transformer} 
The emergence of transformer-based models~\cite{TNNLS_zhang2022adaptive,TNNLS_liu2023survey} in computer vision, initiated by the success of ViT~\cite{dosovitskiy2020image}, has inspired researchers to explore their applicability to video related tasks. TimeSformer~\cite{bertasius2021space} extends ViT to videos by incorporating temporal information directly into the attention mechanism. In MViT~\cite{fan2021multiscale}, a hierarchical architecture is presented, which efficiently captures information across multiple scales, resulting in enhanced performance for a range of vision tasks such as image classification and video understanding. In Video Swin Transformer~\cite{liu2022video}, Swin Transformer~\cite{liu2021swin} is adapted for video understanding tasks, resulting in state-of-the-art performance by proficiently capturing spatiotemporal information within videos. 

In the realm of video segmentation, transformers are instrumental in strengthening the associations of objects in different frames, addressing a previous formidable challenge.
VisTR~\cite{wang2021end} leverages transformers for video instance segmentation by directly processing video frames as a sequence of image patches, thereby capturing both spatial and temporal information effectively. IFC~\cite{hwang2021video} employs inter-frame communication transformers to effectively address video instance segmentation tasks by sharing information between frames.
AOT~\cite{yang2021associating} associates objects across video frames via cross-attention and self-attention, leading to improved performance in semi-supervised video object segmentation. In this work, we design a Mixed Temporal Transformer to enhance temporal consistency. This innovation enable our method to proficiently handle long-term and high-dynamic video sequences, ensuring improved performance and stability in rapidly evolving visual scenes.







\section{Methodology}
In this section, we start by presenting an overview of our proposed MTNet architecture in \S~\ref{sec:overview}. Subsequently, we delve into the details of Bi-modal Fusion Module, Mixed Temporal Transformer, and Cascaded Transformer Decoder in \S~\ref{sec:bfm}, \S~\ref{sec:mtt}, and \S~\ref{sec:ctd}, respectively. Lastly, we furnish the details of loss functions in \S~\ref{sec:loss}.

\subsection{Overview}
\label{sec:overview}
Given an input video $\mathcal{V} = \{V_i \in \mathbb{R}^{w \times h \times 3}\}_{i=1}^N$, the objective of UVOS is to compute binary segmentation masks for the corresponding frames: $\mathcal{S} = \{S_i \in {0,1}^{w \times h}\}_{i=1}^N$. To achieve this, we segment the length of the input video into $C$ clips, with the number of clips $C$ calculated as $\frac{N}{T}$, and $T$ denoting the length of each clip. Following the approach of HFAN~\cite{pei2022hierarchical}, we employ RAFT~\cite{teed2020raft} to extract the optical flow, denoted as $\mathcal{O}=\{O_i\in \mathbb{F}^{w\times h \times 3}\}_{i=1}^N$. Subsequently, the extracted optical flow is partitioned into $C$ clips, in accordance with the division of the input video. 
The overall pipeline of MTNet is shown in Fig.~\ref{figure:pipeline}(a), which primarily consists of three components: Bi-modal Fusion Module, Mixed Temporal Transformer, and Cascaded Transformer Decoder. Initially, ConvNeXt~\cite{liu2022convnet} serves as the universal encoder in our approach, extracting both appearance and motion features from video frame clips and their corresponding flow maps. This extraction process involves four stages, denoted by $k\in [1,2,3,4]$. In all stages of the encoder, Bi-modal Fusion Module is employed to fuse the corresponding appearance and motion features, thereby enhancing the understanding of interdependencies and interactions within multi-modality feature sets. At the last two stages, resolutions are reduced to $\frac{H}{16}\times \frac{W}{16}$ and $\frac{H}{32}\times \frac{W}{32}$, and the Mixed Temporal Transformer is utilized to model temporal relationships between frames efficiently. Finally, the acquired multi-level features are input into the Cascaded Transformer Decoder to generate precise mask predictions for the video clip.

\subsection{Bi-modal Fusion Module}
\label{sec:bfm}
\subsubsection{Multi-modal Gate Unit}
In various video tasks~\cite{zhu2017deep,xue2019video}, optical flow provides crucial motion data between successive frames, enabling the model to learn precise estimations of temporal variations.  
Prior research in unsupervised video object segmentation (UVOS)~\cite{zhou2020motion,yang2021learning,ji2021full,zhang2021deep,pei2022hierarchical} has employed elaborate designs to align optical flow with video frames, guiding the prediction process. While these approaches have led to significant performance improvements, 
their complexity can hinder the efficiency of training and inference processes.
 In light of this, we introduce a more streamlined approach by developing Bi-modal Fusion Modules (BFMs) that efficiently combine features at each level derived from the encoder. We denote the extracted appearance features and motion features as ${\{\mathcal{A}_k\}}_{k=1}^K$ and ${\{\mathcal{M}_k\}}_{k=1}^K$ respectively. For brevity, we describe the process at the $k$-level, though this operation is applicable to other levels as well. As shown in Fig.~\ref{figure:pipeline}, 
The appearance and motion features are initially compressed using two separate 3x3 convolutional layers, followed by a concatenation of the resulting outputs. 
Subsequently, these combined features are processed through a series of operations designed to obtain the weighted vectors for each modality:
\begin{equation}
    \mathcal{F}_k = Conv(Cat(Conv_S(\mathcal{A}_k),Conv_S(\mathcal{M}_k))),
\end{equation}
\begin{equation}
\mathcal{F}_k^A, \mathcal{F}_k^M = Split(\mathcal{F}_k)
\end{equation}
\begin{equation}
    g^A = GAP(\sigma(\mathcal{F}_k^A)),  
\end{equation}
\begin{equation}
    g^M = GAP(\sigma(\mathcal{F}_k^M)), 
\end{equation}


$Conv_S$ and $\sigma$ represent the $1\times1$ convolution used to to shrink feature dimensions and the Sigmoid function, respectively. The fused feature $\mathcal{F}_k\in \mathbb{R}^{H\times W\times 2}$ is obtained by concatenating appearance features $\mathcal{A}_k$ and motion features $\mathcal{M}_k$, which is then processed through a $1\times 1$ convolutional layer for initial fusion. This integrated feature is then divided into two groups, $\mathcal{F}_k^A$ and $\mathcal{F}_k^M$,  each undergoing further processing with Sigmoid and Global Average Pooling (GAP) operations to derive the weighted vector $g^A$ and $g^M$ for each respective modality. Then, the dot production $\odot$ is calculated between input features and the corresponding weighted vector, allowing for an adaptive enhancement of features across both modalities:
\begin{equation}
    \hat{\mathcal{A}}_k = g^A \odot \mathcal{A}_k,
\end{equation}
\begin{equation}
    \hat{\mathcal{M}}_k = g^M \odot \mathcal{M}_k,
\end{equation}
\subsubsection{Co-attention Fusion}
The fused features undergo further concatenation and are processed through a parallel of co-attention operations:
\begin{equation}
    \mathcal{R}_k = Cat(\hat{\mathcal{A}}_k,\hat{\mathcal{M}}_k) 
\end{equation}
\begin{equation}
    \hat{\mathcal{R}}_k = \sigma(C_{attn}(\mathcal{R}_k)+S_{attn}(\mathcal{R}_k))
\end{equation}
$C_{attn}$ and $S_{attn}$ denote co-channel attention and co-spatial attention, respectively. A Sigmoid function is utilized to normalize the resultant values to the interval [0,1], enabling effective re-weighting. These mechanisms are specifically engineered to articulate the interaction between appearance and motion information across channel and spatial dimensions. Specifically, $C_{attn}$, the channel co-attention component, refines the weighting of each channel within the combined multi-modal feature set  $\mathcal{R}_k$, thereby optimizing the integration of these modalities:
\begin{equation}
    \mathcal{R}_k^{avg}= Fc2(\phi (Fc1(Avgpool(\mathcal{R}_k)))
\end{equation}
\begin{equation}
    \mathcal{R}_k^{max}= Fc2(\phi (Fc1(Maxpool(\mathcal{R}_k)))
\end{equation}
\begin{equation}
    C_{attn}(\mathcal{R}_k)= \mathcal{R}_k^{avg} + \mathcal{R}_k^{max}
\end{equation}
$Fc$ is a fully connected layer, and $\phi $ is the ReLU activation function. The terms $Avgpool$ and $Maxpool$ refer to average pooling and max pooling, respectively. For spatial co-attention, $S_{attn}$ generates a spatial attention feature map that effectively captures the global spatial distributions of $\mathcal{R}_k$  through the following process:
\begin{equation}
    S_{attn}(\mathcal{R}_k)= Conv(Cat(Mean(\mathcal{R}_k), Max(\mathcal{R}_k)))
\end{equation}
$Mean$ and $Max$ operations are computed along the channel dimension. 
With the re-weighted $\hat{\mathcal{R}}_k$, we can finally obtain the output of BFM via:
\begin{equation}
    B_k = \hat{\mathcal{R}}_k \odot \hat{\mathcal{A}}_k + (1-\hat{\mathcal{R}}_k)\odot \hat{\mathcal{M}}_k
\end{equation}
$k\in \{1:K\}$ represents the features from different stages of the backbone, in alignment with prior works~\cite{ji2021full,pei2022hierarchical}, we set $K=4$ for our experiments.

\begin{figure}[t]
\centering
\vspace{-2mm}
\begin{tabular}{@{}c}
\includegraphics[width=0.98\linewidth]{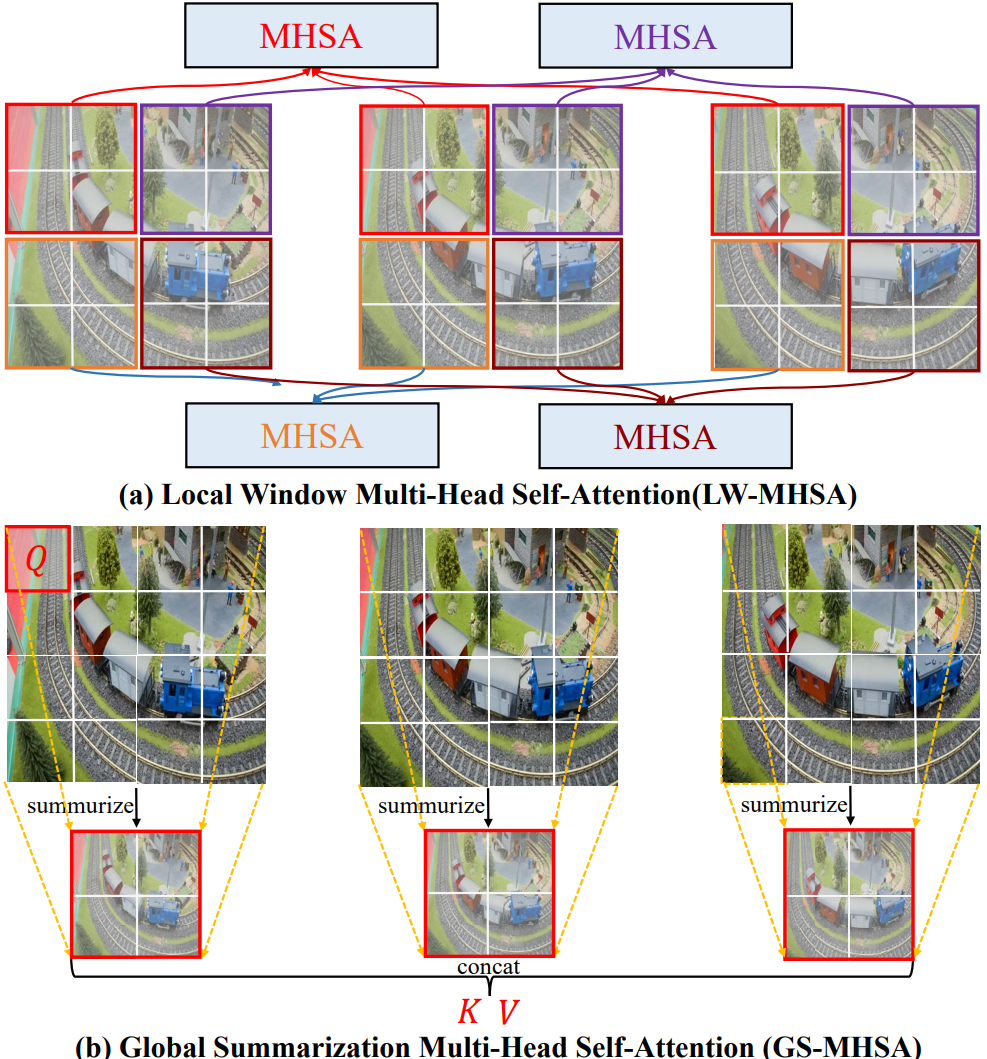} \\
\end{tabular}
\vspace{-1mm}
\caption{Illustration of (a) Local Window MHSA and (b) Global Summarization MHSA.}
\label{figure:LTTF_GTTF}
\vspace{-2mm}
\end{figure} 

\subsection{Mixed Temporal Transformer}
\label{sec:mtt}
With the adaptive multi-modal fusion by BFM, we obtain a compact joint representation for each frame. To enhance the temporal relationships and endow our model with a comprehensive understanding of the video content, we introduce the Mixed Temporal Transformer, which comprises two transformer layers:
the local temporal transformer layer (LTTL) and the global temporal transformer layer (GTTL). This combination effectively captures long-term dependencies in an efficient manner. 
As is shown in Fig.~\ref{figure:pipeline}(b), given the input feature $B_k\in \mathbf{R}^{T\times H\times W\times d}$, where $T$, $H$, $W$, and $d$ represent temporal length, height, width, and dimension, respectively, local temporal transformer layer (LTTL) divides the input feature into $\frac{H\times W}{M^2}$ windows~\cite{liu2021swin}, with each window having dimensions $T\times W\times W\times d$. Within each local window, the standard multi-head self-attention is computed. 
LTTL effectively reduces the computational cost of temporal multi-head self-attention from the original complexity of $O(T^2H^2W^2d)$ to $O(\frac{T^{2}H^{2}W^{2}}{M^{2}}d)$, thus facilitating efficient handling of high-resolution and long-duration videos. However, the receptive field of LTTL is confined within the same location of the temporal window, resulting in limited interactions between different windows. This limitation poses challenges in effectively processing scenes characterized by high dynamics and rapid motion.
To further enhance the dependencies captured by the LTTL and expand its spatiotemporal  receptive field,
we introduced a global temporal transformer layer (GTTL), which reduces the resolution of keys and values in spatiotemporal dimension to achieve computational efficiency when calculating self-attention while meanwhile maintains the global receptive field. The design of the GTTL, partially inspired by the Pyramid Vision Transformer (PVT)~\cite{wang2021pyramid}, aims to minimize computational expenses by compressing the key ($k$) and value ($v$) components in the self-attention mechanism. Specifically, the self-attention in PVT can be formulated as follows:
\begin{equation}
    SRA(Q,K,V)=Attn(Q, SR(K), SR(V))
\end{equation}
\begin{equation}
    Attn(Q,K,V)=Softmax(\frac{QK^{T} }{\sqrt{d } } )V
\end{equation}
$SRA$ denotes spatial-reduction self-attention, and SR refers to the spatial dimension reduction operation of input feature maps, as elaborated in \cite{wang2021pyramid}. In the context of the Global Temporal Transformer Layer (GTTL), we adapt SRA for the video domain. Specifically, for the feature $L_k^{'}$ refined by the Local Temporal Transformer Layer (LTTL), the SR operation is independently applied to each frame's key and value. Subsequently, these modified components are concatenated along the temporal axis to construct video-level key and value matrices. The multi-head self-attention mechanism is then executed using the video-level query $Q_{v}$, key $K_{v}$, and value $V_{v}$, enabling an effective temporal feature integration:
\begin{equation}
    GTTL(Q_{v},K_{v},V_{v})=Attn(Q_{v}, SR(K_{v}), SR(V_{v}))
\end{equation}
Overall, the computational process of the Mixed Temporal Transformer can be represented as:
\begin{equation}
    L_k = LTTL(LN(B_k))+B_k,
\end{equation}
\begin{equation}
    L_k^{'} = FFN(LN(L_k))+L_k,
\end{equation}
\begin{equation}
    G_k = GTTL(LN(L_k^{'}))+L_k^{'},
\end{equation}
\begin{equation}
    G_k^{'} = FFN(LN(G_k))+G_k,
\end{equation}
LN and FFN represents the layer norm and feed forward network in Transformer~\cite{vaswani2017attention}.

Additionally, we present a comparison between Local Window Multi-Head Self-Attention (LW-MHSA) in LTTL and Global Summarization Multi-Head Self-Attention (GS-MHSA) in GTTL in Fig.~\ref{figure:LTTF_GTTF}. This comparison demonstrates that LW-MHSA primarily emphasizes interactions within temporal windows, whereas GS-MHSA extends its reach, capturing a more holistic understanding of the integral information.
By integrating both transformer layers in an interwoven
manner, the Mixed Temporal Transformer adeptly models the relationships between adjacent frames, which is essential for ensuring consistent object localization within the video clip.

\subsection{Cascaded Transformer Decoder}
\label{sec:ctd}
The efficacy of multi-level feature fusion in segmentation tasks has been well-established. In this section, we introduce the Cascaded Transformer Decoder (CTD), a novel architecture designed specifically for multi-level feature fusion and calibration.
Given the input features $\{F_k\}_{k=1}^K$ from four stages, with low-level features $k\in \{1,2\}$ originated from the Bi-modal Fusion Module, and high-level features $k\in \{3,4\}$ derived from Mixed Temporal Transformer, the CTD can regulate the feature, facilitating the transfer of crucial information from deep to shallow layers. Fig.~\ref{figure:pipeline}(a) show the whole decoding process. Specifically, the the whole process in our decoder can be formulated as:
\begin{equation}
    \hat{F}_k = \left\{\begin{matrix}CTD(F_k), \ \ \textup{if} \ k=4
 \\ CTD(F_k, \hat{F_{k+1}}), \ \ \textup{if} \ k=1,2,3
\label{equation:CTD}
\end{matrix}\right.
\end{equation}
Fig.\ref{figure:pipeline}(d) provides a detailed illustration of each Cascaded Transformer Decoder, which draws inspiration from recent advancements in incorporating convolutions into transformer architectures\cite{guo2022cmt,liu2022convnet}. Specifically, we denote the shallower input feature with higher resolution as $F_{shal}$ and the deeper input feature with lower resolution as $F_{deep}$. The Cascaded Transformer Decoder initially extracts information from $F_{shal}$ as follows:
\begin{equation}
\hat{F}_{shal}=CA(DWConv(F_{shal})),
\end{equation}
DWConv signifies the sequential operations involving a depth-wise convolution, followed by batch normalization and a ReLU activation function. CA denotes channel attention. 
The significance of a large receptive field for scene understanding is well-established, as discussed in \cite{ding2022scaling}. Accordingly, we increase the kernel size of DWConv to $7\times7$, enabling more effective capture of the target-object information from deeper layers. Depth-wise convolution, operating independently on each channel, inherently lacks inter-channel interaction. To address this, we incorporate an SE channel attention block, as outlined in \cite{hu2018squeeze}, following the DWConv, to facilitate channel interaction and enhance feature representation.
Subsequently, the deep feature information is incorporated with shallow feature as follows:
\begin{equation}
F = DWConv(Up(F_{deep})+\hat{F}_{shal})+F_{shal},
\end{equation}
\begin{equation}
\hat{F} = FFN(LN(F))+F
\end{equation}
$Up$ refers to bilinear up-sampling, which is employed to align the resolution of features across different scales, and $\hat{F}$ represents the output. 
This design establishes CTD as an efficient block for cross-scale and cross-channel fusion. Through this process, knowledge of the target object, encompassing both spatial and channel dimensions, is progressively propagated from deeper to shallower layers.
Upon completion of the overall process outlined in Eq.\ref{equation:CTD}, 
a $1\times 1$ convolutional layer and bilinear interpolation operation are utilized as the mask decoder to predict the predictions masks $P$.

\subsection{Loss Functions}
\label{sec:loss}
We implement multi-level supervision to enhance the training process by supervising predictions at various levels.
Specifically, the predictions at frame $t$ are represented as $P^t \in  \{P_k^t\}_{k=1}^4$, with $P_1^t$ as the final prediction, and $\{P_s^t\}_{k=2}^4$ as auxiliary predictions derived from various features. Binary cross-entropy loss is used to supervise the training process by comparing $P^t$ and the ground-truth $G$:
\begin{equation}
\begin{split}
        \mathcal{L}=\frac{1}{H\times W}(\sum_{x,y}\mathcal{L}_{BCE}(P_1^t(x,y),G^t(x,y))\\
        +\lambda\sum_{k=2}^4\sum_{x,y}\mathcal{L}_{BCE}(P_k^t(x,y),G^t(x,y))),
\end{split}
\end{equation}

$(x,y)$ represents the spatial coordinates within frame $t$, while $\lambda$ is set to 0.5 to balance loss terms.
During inference and evaluation, we utilize the original prediction results for VSOD, while the 
$\texttt{argmax}$
is further employed to generate binary masks for UVOS.

\section{Experiments}

\subsection{Experimental Setup}
\subsubsection{UVOS Datasets}
We have rigorously evaluated our method across four prominent publicly available datasets, ensuring a comprehensive assessment of its performance. These datasets include DAVIS-16~\cite{perazzi2016benchmark}, FBMS~\cite{ochs2013segmentation}, YouTube-Objects~\cite{prest2012learning}, and Long-Videos~\cite{liang2020video}.

The DAVIS-16 dataset~\cite{perazzi2016benchmark} comprises 50 high-quality videos, meticulously annotated for precise benchmarking. These videos are partitioned into a training set comprising 30 videos and a validation set consisting of 20 videos, ensuring rigorous evaluation standards.

FBMS~\cite{ochs2013segmentation} features 59 videos depicting multiple foreground objects in natural settings. These videos are split into 29 training sequences and 30 test sequences. Consistent with established protocals~\cite{cho2022domain, yang2021learning, zhang2021deep, ren2021reciprocal, zhou2020motion, TNNLS_yin2021directional} in the literature, our model is neither trained nor fine-tuned on the FBMS dataset; instead, we directly evaluate its performance on the 30 designated test video sequences.

The YouTube-Objects dataset~\cite{prest2012learning} is a diverse collection of 126 videos, encompassing 10 distinctive object categories. Notably, this dataset uniquely challenges saliency detection methods by providing ground-truth annotations sparsely at intervals of every ten frames.


The Long-Videos dataset~\cite{liang2020video} is specialized in its focus on longer video content, including three lengthy videos that collectively exceed 7,000 frames. This dataset is particularly valuable for assessing the robustness and consistency of video object segmentation techniques over long durations.



\subsubsection{VSOD Datasets} 
We conduct experiments on four widely-used datasets: DAVIS-16~\cite{perazzi2016benchmark},
ViSal~\cite{wang2015consistent}, SegTrack-V2~\cite{li2013video} and DAVSOD~\cite{fan2019shifting}. DAVIS-16~\cite{perazzi2016benchmark}, commonly used for unsupervised video object segmentation (UVOS), serves as one of our primary evaluation benchmarks. ViSal~\cite{wang2015consistent} and SegTrack-V2~\cite{li2013video} are earlier datasets for video object segmentation, comprising 17 and 13 video sequences respectively. DAVSOD~\cite{fan2019shifting} is distinguished by its complexity, making it a particularly challenging dataset for video segmentation due to its complex scenes, salience shifts, and diverse attributes.

\subsubsection{Evaluation metrics} 
In accordance with~\cite{ji2021full, pei2022hierarchical}, we report mean region similarity ($\mathcal{J}$) and mean boundary accuracy ($\mathcal{F}$) for evaluating UVOS performance. For VSOD, we employ four standard metrics: structure-measure ($S_{\alpha}$, $\alpha$=0.5)~\cite{cheng2021structure}, maximum enhanced alignment measure ($E_{\xi}^{max}$)~\cite{fan2018enhanced}, maximum F-measure ($F_{\beta}^{max}$, ${\beta}^{2}$=0.3)~\cite{achanta2009frequency}, and mean absolute error ($MAE$)~\cite{perazzi2012saliency}.

\subsubsection{Training details} 
All experiments are conducted using the PyTorch Toolkit. Following HFAN~\cite{pei2022hierarchical}, the training procedure is structured in two phases: pre-training on the YouTube-VOS~\cite{xu2018youtube} dataset, followed by fine-tuning on the DAVIS-16~\cite{perazzi2016benchmark} training set. The tiny version of ConvNext~\cite{liu2022convnet} serves as the unified encoder for extracting both appearance and motion features. RAFT is utilized to generate optical flow maps, which are then converted to the three-channel format for compatibility. During training, we adopt the sampling strategy outlined in STCN~\cite{cheng2021rethinking}, selecting three frames from the same video to form a video clip. These video clips are subjected to a variety of data augmentations, including random flips, random crops, random rotations between [-15, 15] degrees, and color jittering; the video order is reversed with a 0.5 probability. All videos are resized to a uniformed resolution of $512 \times 512$. 
We employ the AdamW optimizer and Binary Cross-Entropy (BCE) loss for both stages during training. To enhance efficiency, Automatic Mixed Precision (AMP)~\cite{micikevicius2017mixed} is utilized to accelerate the computational process.

\subsubsection{Inference} Upon finishing the training process, we directly evaluate our model on various datasets without applying any dataset-specific fine-tuning. 
Our model is specifically designed for offline UVOS, capable of handling videos of arbitrary lengths and yielding performance that varies depending on the length. Specifically, for a test video $\mathcal{V}$ and its corresponding flow maps $\mathcal{O}$, each comprising $N$ frame, we first partition $\mathcal{V}$ and $\mathcal{O}$ into $C$ clips, where $C=\lfloor \frac{N}{T} \rfloor$ and $T$ represents the test length of each clip. Subsequently, we feed each clip into our model and directly obtain the clip-level results, enabling simultaneous and seamless primary object segmentation and tracking.


\begin{table*}[t]
\footnotesize
\setlength\tabcolsep{4.73pt}
\renewcommand{\arraystretch}{1.0}
\vspace{-2pt}
\begin{center}
\caption{Quantitative comparison on DAVIS-16 dataset~\cite{perazzi2016benchmark}.  $\uparrow$ ($\downarrow$) denotes that the higher (lower) is better. We use the mean region similarity ($\mathcal{J}$), mean boundary accuracy ($\mathcal{F}$) and $\mathcal{J\&F}$ mean as evaluation metrics. 'PP` denotes post-processing. The top two scores are marked with \textbf{bold} and \underline{underline} respectively.}
\label{tab:davis}
\resizebox{173mm}{!}{
\begin{tabular}{ll|ccc|ccc|c|l}
\toprule[0.9pt]

\multirow{2}{*}{PP\ Public.} & \multirow{2}{*}{Method} &\multicolumn{3}{c|}{$\mathcal{J}$}&\multicolumn{3}{c|}{$\mathcal{F}$}&$\mathcal{J\&F}$&\multirow{2}{*}{FPS}\\

& & $Mean \uparrow$ & $Recall \uparrow$ & $Decay \downarrow $& $Mean \uparrow$ & $Recall \uparrow$ & $Decay \downarrow $ & $Mean \uparrow$ & \\
\midrule[0.65pt]

       
                                       

                                       

\xmark \ \ \   $AAAI_{20}$ ~\cite{gu2020pyramid}   & PCSA
                                   
                                      & 78.1 & 90.0 &	4.4 &	78.5 &	88.1 	&4.1 &	78.3  & \textbf{110} \\

\cmark \ \    $AAAI_{20}$ ~\cite{zhou2020motion}   & MATNet
                                   
                                      &  82.4 & 94.5 &	3.8 &	80.7 &	90.2 & 4.5 & 81.5  &  1.3\\

\xmark \ \ \   $ECCV_{20}$ ~\cite{zhen2020learning}   & DFNet
                                   
                                      &  83.4 &	94.4 &	4.2 & 8.8 &	89.0 &	3.7 & 82.6  & 3.6 \\

\xmark \ \ \   $AAAI_{21}$ ~\cite{liu2021f2net}   & F2Net
                                   
                                      &  83.1 &	95.7 	& \textbf{0.0} &	84.4 &	92.3 & 0.8 &	83.7  & 10.0 \\
                                      
\cmark \ \    $CVPR_{21}$ ~\cite{ren2021reciprocal}   & RTNet
                                   
                                      &  85.6 &	96.1 & - & 84.7 & 93.8  & - & 85.2  & - \\         
                                      
\cmark \ \    $ICCV_{21}$ ~\cite{ji2021full}   & FSNet
                                   
                                      &  83.4 & 94.5 &	3.2 & 83.1 & 90.2 & 2.6  & 84.6  & 12.5 \\

\cmark \ \    $ICCV_{21}$ ~\cite{yang2021learning}   & AMCNet
                                   
                                      &  84.5 & \underline{96.4} &	2.8 & 84.6 & 93.8 &	2.5 & 84.6  & - \\

\cmark \ \    $ICCV_{21}$ ~\cite{zhang2021deep}   & TransportNet
                                   
                                      &  84.5 &	- &	- & 85.0 &	- & - & 84.8  & 3.6 \\
                          
\xmark \ \    $ECCV_{22}$ ~\cite{pei2022hierarchical}   & HFAN
                                   
                                      &  86.8 &   96.1 &  4.3 & 88.2  &  \underline{95.3} & 1.1 & 87.5 & 14.4 \\
                                      

\cmark \ \  $WACV_{23}$ ~\cite{lee2023unsupervised} & PMN
                                       
                                      &  85.9 &  - &  - &  85.6 &  - &  - & 86.2 & - \\

\xmark \ \  $WACV_{23}$ ~\cite{cho2023treating} & TMO
                                       
                                      &  86.1 & - &  - & 85.6 & - &  - & 86.6 & - \\

\cmark \ \    $TIP_{23}$ ~\cite{pei2023hierarchical} & HCPN 
                                       
                                      &   85.8 & 96.2 &  3.4 & 85.4 & 93.2 & 3.0 & 85.6 & 1.2 \\      
\xmark \ \    $ICIP_{23}$ ~\cite{lee2023tsanet} &  TSANet & 86.6 & 95.7 & \textbf{0.0} & 88.3 & 94.3 & \textbf{0.0} & 87.4 & - \\
\cmark \ \    $PR_{24}$ ~\cite{li2024efficient} & LSTA &  82.7 & - & - & 84.8  & - &  - & 83.8 & 36.8 \\          
\cmark \ \    $CVPR_{24}$ ~\cite{lee2024guided} & GSA  &  \underline{88.3} & - &  - & \underline{89.6}  & - & - & \underline{89.0} & - \\                                        
\rowcolor[RGB]{192,192,192}   
\xmark \ \    \textbf{Ours} & \textbf{MTNet} 
                                       
                                      &  \textbf{88.7} &   \textbf{96.9} &  4.2 &  \textbf{90.7} &  \textbf{96.0} &  2.9 & \textbf{89.7} & \underline{43.4} \\
                                       

\bottomrule[0.9pt]
\end{tabular}}
\end{center}

\vspace{-2mm}

\end{table*}

\begin{table*}[t]
\footnotesize
\setlength\tabcolsep{4.73pt}
\renewcommand{\arraystretch}{1.0}
\begin{center}
\caption{Quantitative comparison on YouTube-Objects~\cite{prest2012learning}. We compare the region similarity($\mathcal{J}$) for each category, as well as the overall average score across categories($\mathcal{J}$ Mean), with the top two performances denoted by \textbf{bold} and \underline{underline} respectively.}
\label{tab:youtube}
\resizebox{160mm}{!}{
\begin{tabular}{l|cccccccccc|c}
\toprule[0.9pt]

Method & \textit{Aeroplane} & \textit{Bird} & \textit{Boat} & \textit{Car} & \textit{Cat} & \textit{Cow} & \textit{Dog} & \textit{Horse} & \textit{Motorbike} & \textit{Train} & \textit{Average}\\
\midrule[0.65pt]

       
FST~\cite{papazoglou2013fast} &70.9 &	70.6 &	57.8 &	33.9 &	30.5 &	41.8 &	36.8 &	44.3 &	48.9 & 	39.2 &	46.2 \\
LVO~\cite{tokmakov2017learning} &\underline{86.2} &	81.0 &	68.5 &	69.3 &	58.8 &	68.5 &	61.7 &	53.9 &	60.8 & 	\underline{66.3} &	67.2 \\
PDB~\cite{song2018pyramid}	 &78.0  & 	80.0  &	58.9  &	76.5  &	63.0  &	64.1  &	70.1  &	67.6  &	58.4  &	35.3  &	65.4 \\
AGS~\cite{wang2019learning}	 & \textbf{87.7}  &	76.7  &	72.2  &	78.6  &	69.2  &	64.6  &	73.3  &	64.4  &	62.1  &	48.2  &	69.7 \\
COSNet~\cite{lu2019see} & 81.1  & 75.7  & \underline{71.3}  & 77.6  & 66.5  & 69.8  & 76.8  & 67.4  & \underline{67.7}  & 46.8  & 70.5 \\
AGNN~\cite{wang2019zero} &	81.1  &	75.9  &	70.7  &	78.1  &	67.9  &	69.7  &	77.4  &	67.3  &	\textbf{68.3}  &	47.8  &	70.8 \\
MATNet~\cite{zhou2020motion} &	72.9  &	77.5  &	66.9  &	79.0  &	73.7  &	67.4  &	75.9  &	63.2  &	62.6  &	51.0  &	69.0 \\
RTNet~\cite{ren2021reciprocal} &	84.1  &	80.2  &	70.0  &	79.5  &	71.8 &	70.1  &	71.3  &	65.1 &	64.6  &	53.3  &	71.0 \\
AMC-Net~\cite{yang2021learning} &	78.9  &	80.9  &	67.4  &	\underline{82.0}  &	69.0  &	69.6  &	75.8  &	63.0  &	63.4  &	57.8  &	71.1 \\
HFAN~\cite{pei2022hierarchical} &	84.7  &	80.0  &	\textbf{72.0}  &	76.1  &	76.0  &	\underline{71.2}  &	76.9  &	\textbf{71.0}  &	64.3  &	61.4  &	\underline{73.4} \\
HCPN~\cite{pei2023hierarchical} &	84.5  &	\underline{79.6}  &	67.3  &	87.8  &	\underline{74.1}  &	71.2  &	\underline{76.5}  & \underline{66.2}  &	65.8  &	59.7  & 73.3 \\
                                      
\rowcolor[RGB]{192,192,192}                                      
\textbf{MTNet} & 83.7 &	\textbf{85.5} &	63.5 &	\textbf{83.6} &	\textbf{79.8} &	\textbf{72.6} &	\textbf{81.4} &	67.3 &	56.0 &	\textbf{72.3} &	\textbf{74.6}  \\

\bottomrule[0.9pt]                               

\end{tabular}}
\end{center}


\end{table*}

\begin{table*}[t]
\footnotesize
\setlength\tabcolsep{4.73pt}
\renewcommand{\arraystretch}{1.0}
\vspace{-4pt}
\begin{center}
\caption{Quantitative Evaluation on the FBMS Dataset~\cite{ochs2013segmentation}: Performance Metrics Using $\mathcal{J}$ Mean. The highest and second-highest scoring methods are highlighted in \textbf{bold} and \underline{underline} respectively.}

\label{tab:fbms}
\vspace{-2mm}
\resizebox{180mm}{!}{
\begin{tabular}{l|cccccccccccc >{\columncolor[RGB]{192,192,192}}c}
\toprule[0.9pt]

Method & FST~\cite{papazoglou2013fast} & MSTP~\cite{hu2018unsupervised} & IET~\cite{li2018instance} & PDB~\cite{song2018pyramid} & COSNet~\cite{lu2019see} & MATNet~\cite{zhou2020motion} & F2Net~\cite{liu2021f2net} & MMAPS~\cite{zhao2021multi} & AMC-Net~\cite{yang2021learning} & TransportNet~\cite{zhang2021deep} & EFS~\cite{lee2022iteratively} & TMO~\cite{cho2023treating} & Ours\\
\midrule[0.65pt]


$\mathcal{J} Mean \uparrow$ & 55.5 &	60.8 &	71.9 &	74.0 &	75.6 &	76.1 &	77.5 &	76.7 &  76.5 &	78.7 &	77.5 &  \underline{79.9} & \textbf{83.8}  \\
\bottomrule[0.9pt]                               

\end{tabular}}
\end{center}

\vspace{-4mm}

\end{table*}

\subsection{Quantitative Comparisons with State-of-the-Art Models}
We show the performance comparisons of our MTNet with other state-of-the-art methods on three UVOS benchmarks and four VSOD benchmarks.

\subsubsection{DAVIS-16} 
The DAVIS-16 validation set, comprising 20 videos, is the most commonly used benchmark in Unsupervised Video Object Segmentation(UVOS). Our experiment includes a detailed comparison with 18 state-of-the-art algorithms published between 2019 and 2023, including 
AGNN~\cite{wang2019zero}, COSNet~\cite{lu2019see}, 
PCSA~\cite{gu2020pyramid}, MATNet~\cite{zhou2020motion}, DFNet~\cite{zhen2020learning}, F2Net~\cite{liu2021f2net}, RTNet~\cite{ren2021reciprocal}, FSNet~\cite{ji2021full}, AMCNet~\cite{yang2021learning}, TransportNet~\cite{zhang2021deep}, HFAN~\cite{pei2022hierarchical}, PMN~\cite{lee2023unsupervised}, TMO~\cite{cho2023treating},  HCPN~\cite{pei2023hierarchical}, TSANet~\cite{lee2023tsanet}, LSTA~\cite{li2024efficient}, GSA~\cite{lee2024guided}.
As illustrated in Tab.~\ref{tab:davis}, MTNet surpasses the recent state-of-the-art method GSA~\cite{lee2024guided} in terms of $\mathcal{J}$ Mean, $\mathcal{F}$ Mean, and $\mathcal{J\&F}$ Mean. Compared to methods~\cite{yang2021learning,ji2021full,pei2022hierarchical,pei2023hierarchical,lee2024guided} 
that rely on  Conditional Random Fields (CRF) or Multi-Scale testing (MS),  our approach demonstrates superior performance in both inference speed and segmentation quality.



\subsubsection{YouTube-Objects} 
Validation experiments on the Youtube-Objects dataset~\cite{prest2012learning} do not require external fine-tuning, thereby providing a robust test of the model's generalization ability across diverse scenarios. As shown in Tab.~\ref{tab:youtube}, our method is outperformed by other approaches in certain categories, including \textit{Aeroplane}, \textit{Boat}, \textit{Horse}, and \textit{Motorbike}. Nevertheless,  our method exhibits superior performance in the majority of categories, as well as in the overall average results. Notably, it excels particularly in the categories of \textit{Dog} and \textit{Train}, surpassing the second-best method by $4.4\%$ and $12.4\%$, respectively. 

\subsubsection{FBMS}
FBMS is a challenging dataset with low-resolution, multiple and small target foreground objects. Our MTNet, when compared against other state-of-the-art methods on this dataset (as demonstrated in Table~\ref{tab:fbms}), exhibits superior performance, surpassing the nearest competitor by a notable margin of 3.9 in $\mathcal{J}$ Mean. This demonstrates MTNet's robust capability to detect, track, and segment multiple target objects in complex video scenes.

\subsubsection{Long-Videos}
Long-term videos present substantial challenges yet are pivotal in tracking tasks~\cite{dai2020high,zhang2021learning}, warranting increased attention in unsupervised video object segmentation. The Long-Videos dataset~\cite{liang2020video} comprises three video sequences, each with an average of 2500 frames. In this benchmark, we compare our method not only with other UVOS approaches but also with more competitive SVOS methods that incorporate additional priors. As shown in Tab.~\ref{tab:longvideo}, the proposed MTNet surpasses HFAN~\cite{pei2022hierarchical} by $1.2\%$ in $\mathcal{J\&F}$ Mean, highlighting the effectiveness of the temporal modules in handling long-term videos. Nevertheless, there remains a performance disparity when compared with advanced SVOS methods, indicating areas for further improvement.

\begin{table*}[t]
\footnotesize
\setlength\tabcolsep{4.73pt}
\renewcommand{\arraystretch}{1.0}
\vspace{-2pt}
\begin{center}
\caption{Quantitative comparison on Long-Videos dataset.  The best SVOS and UVOS results are marked with \underline{underline} and \textbf{bold} respectively.}
\label{tab:longvideo}
\vspace{-2pt}
\resizebox{160mm}{!}{
\begin{tabular}{ll|ccc|ccc|c}
\toprule[0.9pt]

\multirow{2}{*}{Public.} & \multirow{2}{*}{Method} &\multicolumn{3}{c|}{$\mathcal{J}$}&\multicolumn{3}{c|}{$\mathcal{F}$}&$\mathcal{J\&F}$\\

& & $Mean \uparrow$ & $Recall \uparrow$ & $Decay \downarrow $& $Mean \uparrow$ & $Recall \uparrow$ & $Decay \downarrow $ & $Mean \uparrow$\\
\midrule[0.65pt]

\multicolumn{4}{l}{\textbf{Performance of SVOS Methods}} \\

$ICCV_{19}$ ~\cite{gu2020pyramid}   & STM
                                   
                                      & 78.1 & 90.0 &	\underline{4.4} &	78.5 &	88.1 	&\underline{4.1} &	78.3 \\

$NIPS_{20}$ ~\cite{zhou2020motion}   & AFB-URR
                                   
                                      &  82.7 &	\underline{91.7} 	& 11.5 &	83.8 &	\underline{91.7} &	13.9 &	83.3  \\

$NIPS_{21}$ ~\cite{zhen2020learning}   & AOT
                                   
                                      &  \underline{83.2} & - & - & \underline{85.4} & - & -	& \underline{84.3}  \\
                                   
\midrule[0.65pt]

\multicolumn{4}{l}{\textbf{Performance of UVOS Methods}} \\

$AAAI_{20}$ ~\cite{gu2020pyramid}   & AGNN	&	68.3 &	77.2 &	13.0 &	68.6 &	77.2 &	16.6 &	68.5  \\

$AAAI_{20}$ ~\cite{zhou2020motion}   & MATNet	& 66.4 & 73.7 &	10.9 &	69.3 &	77.2 &	10.6 &	67.9  \\

$ECCV_{22}$ ~\cite{zhen2020learning}   & HFAN & 80.2 & 	91.2 & 	\textbf{9.4} & 	83.2 & 	96.5 &  	7.1 & 	81.7  \\
                                      
\rowcolor[RGB]{192,192,192}                                       
\textbf{Ours} & \textbf{MTNet} 
                                       
                                      &  \textbf{79.6} &   \textbf{91.2} &  9.5 &  \textbf{85.8} &  \textbf{96.7} &  \textbf{6.7} & \textbf{82.7} \\
                                       

\bottomrule[0.9pt]
\end{tabular}
}
\end{center}

\vspace{-5mm}

\end{table*}
\begin{table*}[t]
\footnotesize
\setlength\tabcolsep{4.73pt}
\renewcommand{\arraystretch}{1.1}
\begin{center}
\caption{Quantitative comparison on benchmark VSOD datasets.  $\uparrow$ ($\downarrow$) denotes that the higher (lower) is better. We use the Mean Absolute Error ($MAE$), max F-measure ($F_m^{max}$), S-measure ($S_m$), and max E-measure ($E_m^{max}$) as evaluation metrics. \textbf{Bold} and \underline{underline} denotes the best and secondary results. * indicates that the reported results were obtained through our own re-measurement process. It is pertinent to note that MTNet is primarily trained on the DAVIS-16.$\dag$indicates the use of training datasets similar to CoSTFormer, specifically including both DAVIS-16 and DAVSOD.}
\label{tab:vsod1}
\resizebox{176mm}{!}{
\begin{tabular}{ll|llll|llll|llll|llll}
\toprule[0.9pt]

\multirow{2}{*}{Public.} & Dataset &\multicolumn{4}{c|}{DAVIS-16~\cite{perazzi2016benchmark}} &  \multicolumn{4}{c|}{ViSal~\cite{wang2015consistent}} & \multicolumn{4}{c|}{SegTrack-V2~\cite{li2013video}} & \multicolumn{4}{c}{DAVSOD~\cite{fan2019shifting}}\\
\cline{3-6} \cline{7-10} \cline{11-14} \cline{15-18}

& Metric & 
        $S_{m}\uparrow$ &  $E_m\uparrow $ & $F_m\uparrow$ & $MAE\downarrow$ &
        $S_{m}\uparrow$ &  $E_m\uparrow $ & $F_m\uparrow$ & $MAE\downarrow$ &
        $S_{m}\uparrow$ &  $E_m\uparrow $ & $F_m\uparrow$ & $MAE\downarrow$ &
        $S_{m}\uparrow$ &  $E_m\uparrow $ & $F_m\uparrow$ & $MAE\downarrow$ \\
        
\midrule[0.65pt]
$ECCV_{18}$ ~\cite{song2018pyramid} & PDB
                                      &  0.882 & $0.951^*$ & 0.855 & 0.028 &
                                      0.907 & - & 0.888 & 0.032
                                      & 0.864 & - &	0.800 &	0.024 
                                      & 0.698 & - &	0.572 &	0.116 \\
$CVPR_{19}$ ~\cite{fan2019shifting} & SSAV
                                      &  0.893 & 0.948	& 0.861	& 0.028	&
                                      0.943	& $0.977^*$	& 0.939	& 0.020	&
                                      0.851 &	$0.918^*$ &	0.801 &	0.023 & 
                                      0.755 & 0.806 &	0.659 &	0.084 \\            

$ICCV_{19}$ ~\cite{yan2019semi} & RCR
                                       
                                      &  0.886 &	0.947 & 0.848	& 0.027	& 
                                      0.922	& $0.955^*$	& 0.906	& 0.026	&
                                      $0.843^*$	& $0.878^*$	& $0.780^*$	& $0.035^*$ & 
                                      0.741 & 0.083 &	0.653 &	0.087 \\   

$AAAI_{20}$ ~\cite{gu2020pyramid} & PSCA
                                       
                                      &  0.902 & 0.961 & 0.880 & 0.022 &
                                      0.946 &	$0.983^*$ &	0.940 &	0.017  &
                                      0.865 & 0.907 &	0.810 & 0.025 &
                                      0.741 & 0.793 &	0.656 &	0.086 \\   
                                      
$ICCV_{21}$ ~\cite{zhang2021dynamic} & DCFNet
                                       
                                      &  0.914 &	$0.969^*$ &	0.900 &	0.016	&
                                      0.952	& \underline{$0.990^*$}	& 0.953 &  \textbf{0.010} &	
                                      0.883 &	$0.935^*$ &	\underline{0.839} &	0.015 &
                                      0.741 & $0.805^*$ &	0.660 &	0.074\\

                                       

                                      

$ICCV_{21}$ ~\cite{ji2021full} & FSNet
                                       
                                      &  0.920	& 0.970	& 0.907	& 0.020 & 
                                      $0.923^*$	& $0.972^*$	& $0.907^*$	& $0.023^*$	& 
                                      $0.871^*$	& $0.910^*$	& $0.804^*$	& $0.024^*$	& 
                                       0.773 & \underline{0.825} &	0.685 &	0.072 \\

$ECCV_{22}$ ~\cite{pei2022hierarchical} & HFAN           
                                      & 0.938 & 0.983 & \underline{0.935} & 
                                      \textbf{0.008} &	
                                      $0.944^*$	& $0.980^*$	& $0.927^*$	& $0.014^*$	& 
                                      $0.876^*$ & $0.951^*$ & $0.828^*$ & $0.015^*$  & $0.754^*$ & $0.788^*$ &	$0.645^*$ &	$0.077^*$\\

$TNNLS_{23}$ ~\cite{TNNLS_liu2023learning} & CoSTFormer           
                                      & 0.926 & - & 0.916 & 0.013 & - & - & -	& -	& 
                                     0.888 & - & 0.833 & 0.015  & \underline{0.783} & - & \underline{0.697} & 0.068 \\

\rowcolor[RGB]{192,192,192}                               
                                \textbf{Ours} & \textbf{MTNet} & \textbf{0.951}	&  \underline{0.987} &  \underline{0.944} & \underline{0.010} &	 \textbf{0.959} & \textbf{0.991} &	 \textbf{0.959} & \underline{0.011} &  \underline{0.899} & \underline{0.962} &  \underline{0.875} & \underline{0.013} & 
                                0.780 & \underline{0.825} &	0.688 &	\underline{0.067}\\

\rowcolor[RGB]{192,192,192}                               
                                \textbf{Ours} & \textbf{MTNet}$^{\dag}$ & \underline{0.950}	&  \textbf{0.989} &  \textbf{0.946} & \underline{0.010} & \underline{0.956} & \textbf{0.989} & \underline{0.957} & 0.012 &  \textbf{0.923} & \textbf{0.980} & \textbf{0.904} & \textbf{0.012} & 
                                \textbf{0.820} & \textbf{0.868} &	\textbf{0.750} &	\textbf{0.060}\\
                                
\bottomrule[0.9pt]                           
\end{tabular}
}
\end{center}

\vspace{-2mm}

\end{table*}
\begin{figure*}[!tb]
    \centering
    \vspace{-1mm}
    \includegraphics[scale=0.58] {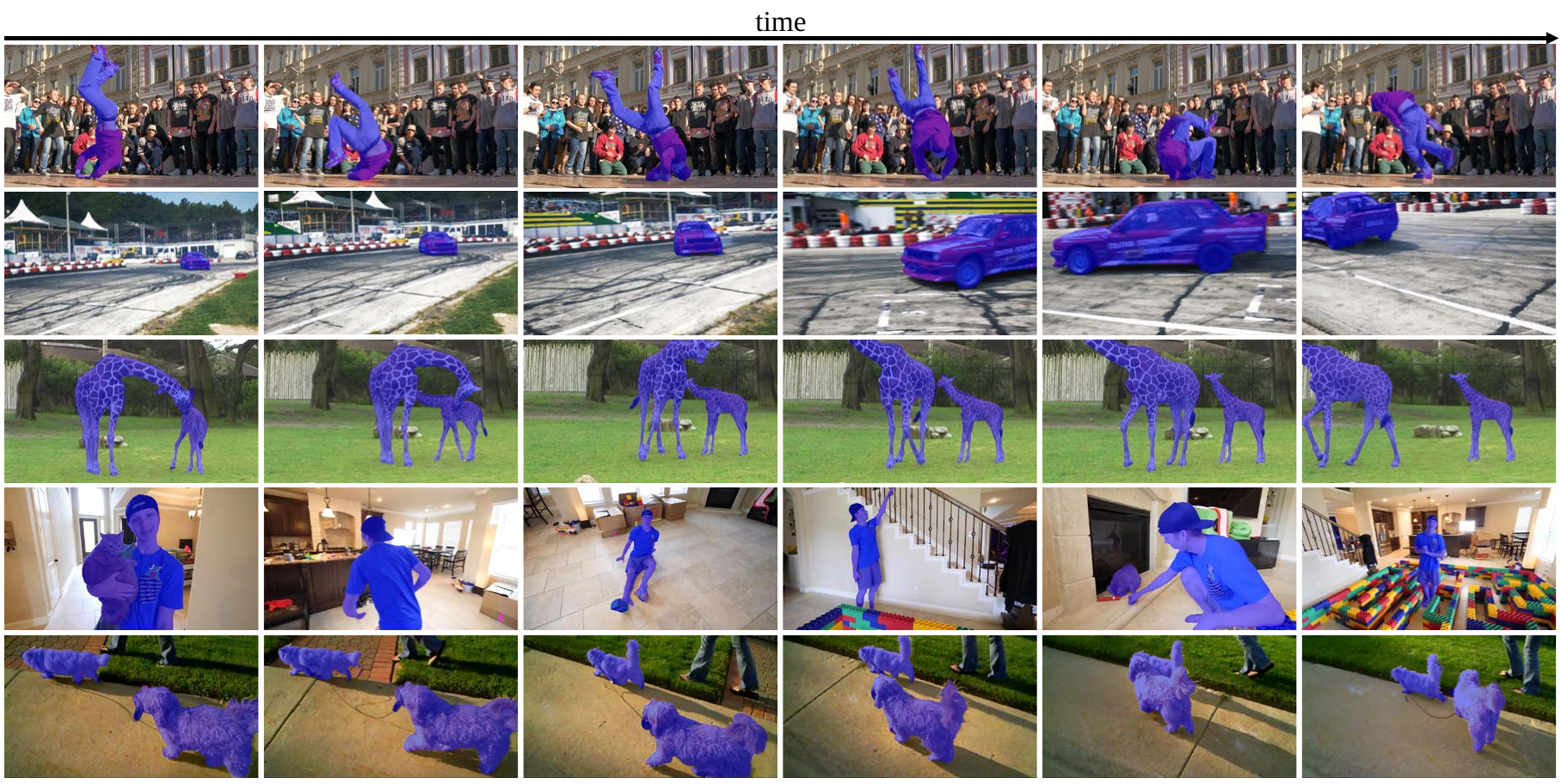} 
    \vspace{-2mm}
    \caption{Visual analysis of unsupervised video object segmentation performance across a variety of video scenarios. The sequence of frames illustrates the algorithm's ability to maintain consistent object segmentation over time, highlighting its effectiveness in various contexts, from group gatherings to fast-moving vehicles and interactions in natural settings.}
    \vspace{-3mm}
    \label{Fig:Visual}
\end{figure*}

\subsubsection{Main results on VSOD datasets} 
We compare the performance of our MTNet in four VSOD datasets, \ie, DAVIS-16~\cite{perazzi2016benchmark}, ViSal~\cite{wang2015consistent}, SegTrack-V2~\cite{li2013video} and DAVSOD~\cite{fan2019shifting}. 
Specifically, we compare MTNet with 7 state-of-the-art VSOD methods. The results are sourced either directly from the original publications or re-measured by us, adhering to a strict evaluation process that employs the original testing codes and model weights in their projects. Notably, to maintain consistency with the training protocol employed by CoSTFormer, the most recent VSOD-specific method, we also adopt a joint training strategy using both DAVIS-16 and DAVSOD datasets. As shown in Tab.~\ref{tab:vsod1}, our proposed MTNet demonstrates superior performance in the majority of datasets, consistently achieving the best results across various evaluation metrics. In DAVIS-16, MTNet surpasses the previously top-performing model, HFAN~\cite{pei2022hierarchical} by 1.4\% and 1.0\% in terms of $S_m$ and $F_m$ respectively, and similar leading trend can also be observed in ViSal, SegTrack-V2 and DAVSOD. 

\begin{figure*}[!tb]
    \centering
    \vspace{-1mm}
    \includegraphics[scale=0.30] {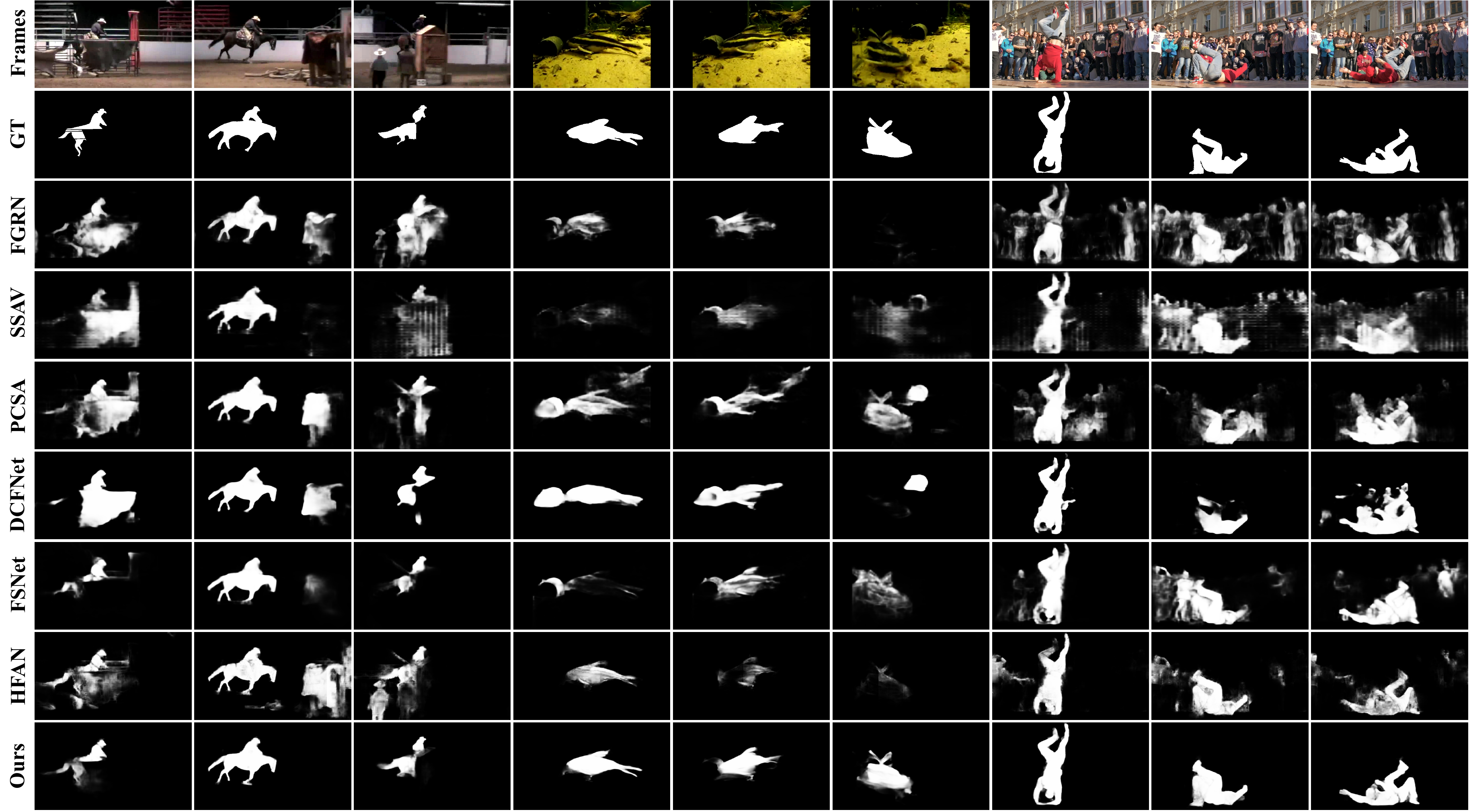} 
    \vspace{-3mm}
    \caption{Visual comparison of the saliency maps between our methods and state-of-the-art models. This figure illustrates the precision and clarity with which our method delineates salient objects, maintaining consistency and accuracy against ground truth (GT) benchmarks. The saliency maps demonstrate our method's superior performance in detecting and segmenting salient objects in diverse and dynamic video scenes.}
    \vspace{-3mm}
    \label{Fig:Visual_VSOD}
\end{figure*}

\subsection{Visual Results}
\subsubsection{Visualization Results of UVOS} 
We present visualization results of our method 
in Fig.~\ref{Fig:Visual}, illustrating the capability of our approach to produce high-quality outcomes. In the 1$st$ row, the person in red shirt and blue jeans is engaged in street dance with continuously changing poses. Our MTNet could consistently captures the person despite the background is full of onlookers, which is quite distracting.
Similar phenomena can be observed in 2$nd$ and 4$th$ rows as well, characterized by the dramatic transformations in the scale and appearance of the racing car undergo dramatic changes, and the extensive mobility of  `blueboy' throughout the room. In  scenarios involving multiple co-existing objects (3$rd$ and 5$th$ rows), our method consistently achieves accurate tracking and segmentation. These cases demonstrate the robustness of our approach in effectively distinguishing the target subject from complex and dynamic backgrounds.

\subsubsection{Visualization Comparisons of VSOD} 
For a clearer comparison of our method against other state-of-the-art techniques in the VSOD task, we present a visual analysis in Fig.~\ref{Fig:Visual_VSOD}. Three distinct scenes, each containing three consecutive frames, are chosen for comparison against six different methods. In the first scene, our method adeptly handles rapid changes in object location and appearance, where many other methods struggle. The second scene features objects so subtle that they challenge even human identification, yet our technique successfully detects motion shifts and precisely segment the primary object. The third scenario showcases a street dancer performing dynamic movements against the crowd background. In this context, our method accurately delineates the dancer, unlike other approaches that struggle to maintain precise segmentation.

\begin{figure}[t]
\centering
\begin{tabular}{@{}c}
\includegraphics[width=0.96\linewidth]{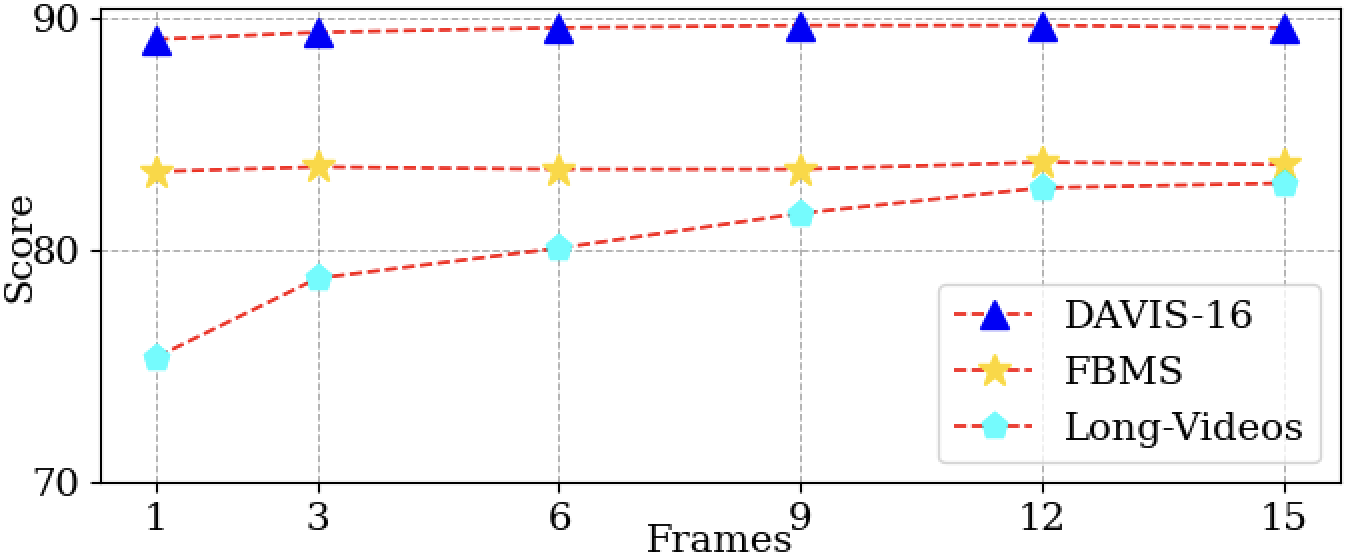} \\
\end{tabular}
\caption{Impact of Clip Length on $\mathcal{J\&F}$ Mean Score Across Datasets. The line chart shows an improvement trend with longer clips in Long-Videos, contrasting with the negligible changes in DAVIS-16 and FBMS, potentially due to their less complex scenarios, which typically feature more distinct objects and less dramatic long-term variations.}
\label{figure:frames}
\end{figure}

\begin{figure}[t]
\vspace{-2mm}
\centering
\begin{tabular}{@{}c}
\includegraphics[width=0.80\linewidth]{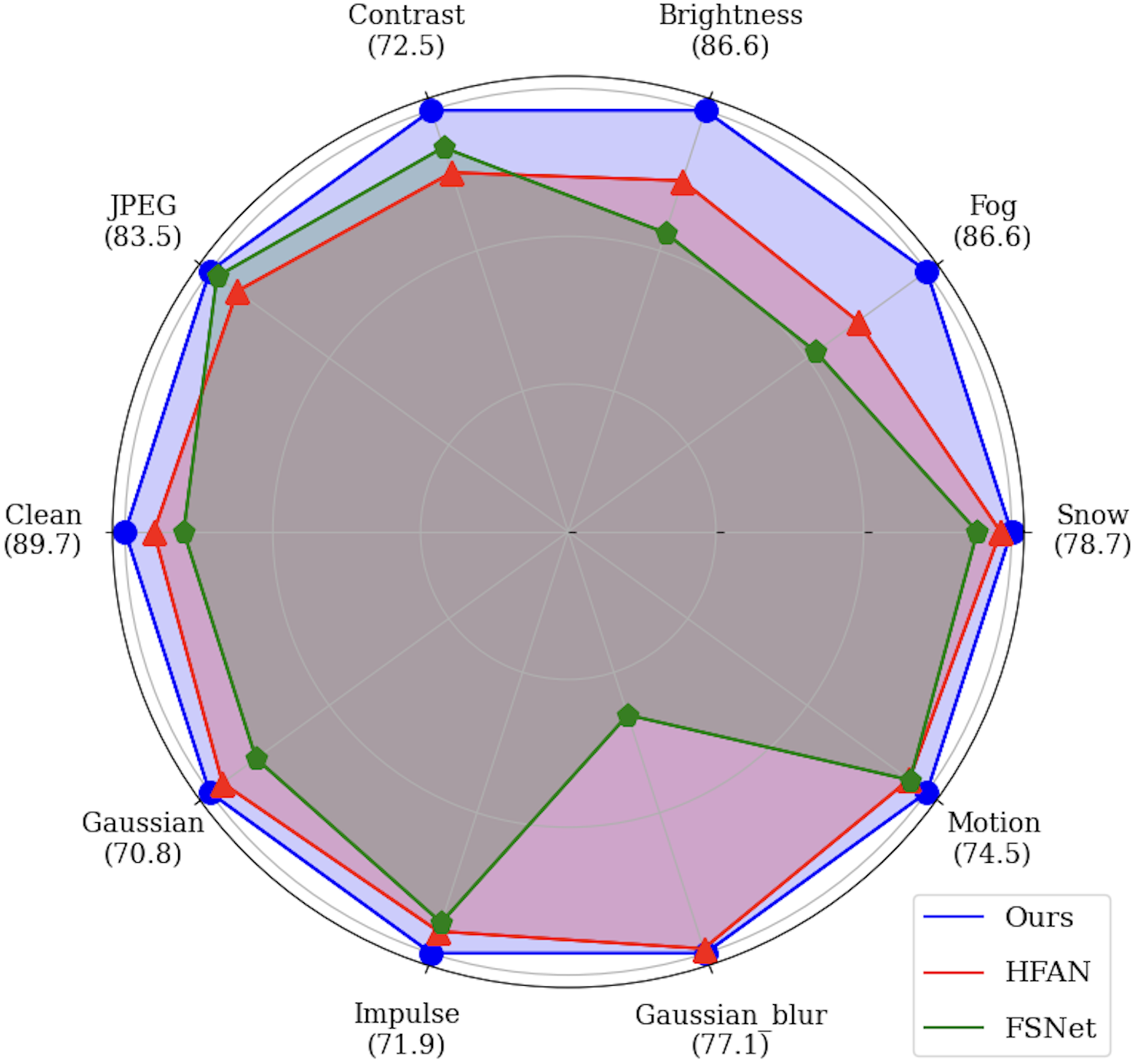} 
\end{tabular}
\caption{$\mathcal{J\&F}$ Mean when applying various corruptions on DAVIS-16~\cite{perazzi2016benchmark}.}
\label{figure:robustness}
\vspace{-2mm}
\end{figure}


\subsection{Ablation Studies}
In this section, we conduct ablation studies on the DAVIS-16 and FBMS datasets. We use the ConvNext version of MTNet as the standard model in our ablation study, with the results reported in Tab.~\ref{tab:ablation} (\#1).

\subsubsection{Design choices of modules} 
We introduce the Bi-modal Fusion Module (BFM), Mixed Temporal Transformer (MTT), and Cascaded Transformer Decoder (CTD) to perform robust and accurate object tracking and segmentation. To demonstrate the efficacy of each component, we conducted a series of validation experiments. Tab.~\ref{tab:ablation} (\#2) presents the results of our baseline, which is a simple combination of the encoder and FPN without any specific designs. Experiments (\#3-\#7) add different components to the baseline in a controlled manner. Compared with the baseline results, it is evident that each component significantly enhances performance. In particular, adding either MTT or CTD results in a noticeable gain, with both components increasing the average $\mathcal{J\&F}$ Mean by 3.6. When both components are used, the absolute increase further reaches 5.5, yielding the most effective performance across all tested configurations.

\subsubsection{Impact of modality} To investigate the influence of input modality on performance, we conduct experiments and present the results in Tab.~\ref{tab:ablation} (\#8-\#11). In these experiments, we employ a single-stream encoder for feature extraction, and the Bi-modal Fusion Module is intentionally excluded.  The term \textit{input} denotes the single modality input mode. The terms \textit{`Add'} and \textit{`Concat'} involve pre-fusion of images and flow maps either through addition or concatenation, respectively, prior to input into the model. 
The observed decline in performance for both modifications underscores the importance of utilizing both modalities and the dual-branch architecture.

\begin{table}[t]
\footnotesize
\setlength\tabcolsep{4.73pt}
\renewcommand{\arraystretch}{1.0}
\begin{center}
\caption{Ablation Study on DAVIS-16~\cite{perazzi2016benchmark} and FBMS~\cite{ochs2013segmentation}: Comparative Analysis using Arabic Numerals for Various Settings including Proposed Method, Module Ablations, Input Modality, and Training Process. $\bigtriangleup$ represents the average performance deviation from MTNet across benchmarks.}
\label{tab:ablation}
\resizebox{70mm}{!}{
\begin{tabular}{l|c|cc|c}
\toprule[0.9pt]
\# & Method & DAVIS-16 & FBMS & $\bigtriangleup$\\
\midrule[0.65pt]
1&MTNet&89.7&83.8&- \\
2&Baseline&86.8&75.9&\textbf{-5.5} \\
3&$w/$ BFM &87.6&77.0&\textbf{-4.5} \\
4&$w/$ MTT &88.7&81.2&\textbf{-1.9} \\
5&$w/$ CTD &88.9&81.0&\textbf{-1.9} \\
6&$w/$ BFM+MTT &89.2&80.5&\textbf{-2.0} \\
7&$w/$ BFM+CTD &89.3&82.1&\textbf{-1.1} \\
\midrule[0.65pt]
8& $Input$ Appearance&85.9&79.2&\textbf{-4.3} \\
9& $Input$ Motion&82.0&63.1&\textbf{-16.0} \\
10& \textit{Add} Motion\&Appearance&87.1&72.3&\textbf{-7.1} \\
11& \textit{Concat} Motion\&Appearance&86.3&77.2&\textbf{-5.0} \\
\midrule[0.65pt]
12& $w/o$ Fine-tuning &82.7&71.2&\textbf{-9.9} \\
13& $w/o$ Pre-training &84.8&75.2&\textbf{-6.8} \\

\bottomrule[0.9pt]
\end{tabular}}
\end{center}

\vspace{-3.5mm}

\end{table}

\subsubsection{Influence of training stages} Since our method undergoes a two-stage training process, \ie, pre-training on YouTube-VOS~\cite{xu2018youtube} first and then fine-tuning on DAVIS-16~\cite{perazzi2016benchmark}, we conduct ablation experiments to evaluate the impact of each stage. The results can be seen in Tab.~\ref{tab:ablation} (\#12,\#13). When removing the fine-tuning stage and pre-training stage, the model experiences significant performance decreases of 9.9 and 6.8, respectively, indicating that both training stages are crucial.

\subsubsection{Clip length} We examine the impact of varying clip lengths during the inference stage. Our experiments are conducted on three benchmark datasets: DAVIS-16~\cite{perazzi2016benchmark}, FBMS~\cite{ochs2013segmentation}, and Long-Videos~\cite{liang2020video}. As illustrated in Fig.~\ref{figure:frames}, the $\mathcal{J\&F}$ Mean for DAVIS-16 and FBMS exhibits minimal variation with increasing clip lengths. In contrast, performance on the Long-Videos dataset significantly improves as the clip length increases, with this trend persisting until saturation is reached as $t$ extends. Based on these findings, we set $t=12$ as the standard clip length in our experiments.

\subsubsection{Robustness to corruptions} Robustness is a crucial aspect in various domains, including segmentation and autonomous driving. To evaluate the robustness of our approach, we sample nine common corruptions from ImageNet-C~\cite{hendrycks2019benchmarking} and apply the most intense degrees to the DAVIS-16~\cite{perazzi2016benchmark} validation set. Importantly, all results are obtained through zero-shot testing, without any fine-tuning. As depicted in Fig.~\ref{figure:robustness}, our method consistently exhibits superior robustness compared to other approaches under a diverse range of corruptions. This finding underscores the potential of our method to deliver reliable performance in challenging and variable conditions.

\begin{figure}[t]
\centering
\begin{tabular}{@{}c}
\includegraphics[width=0.92\linewidth]{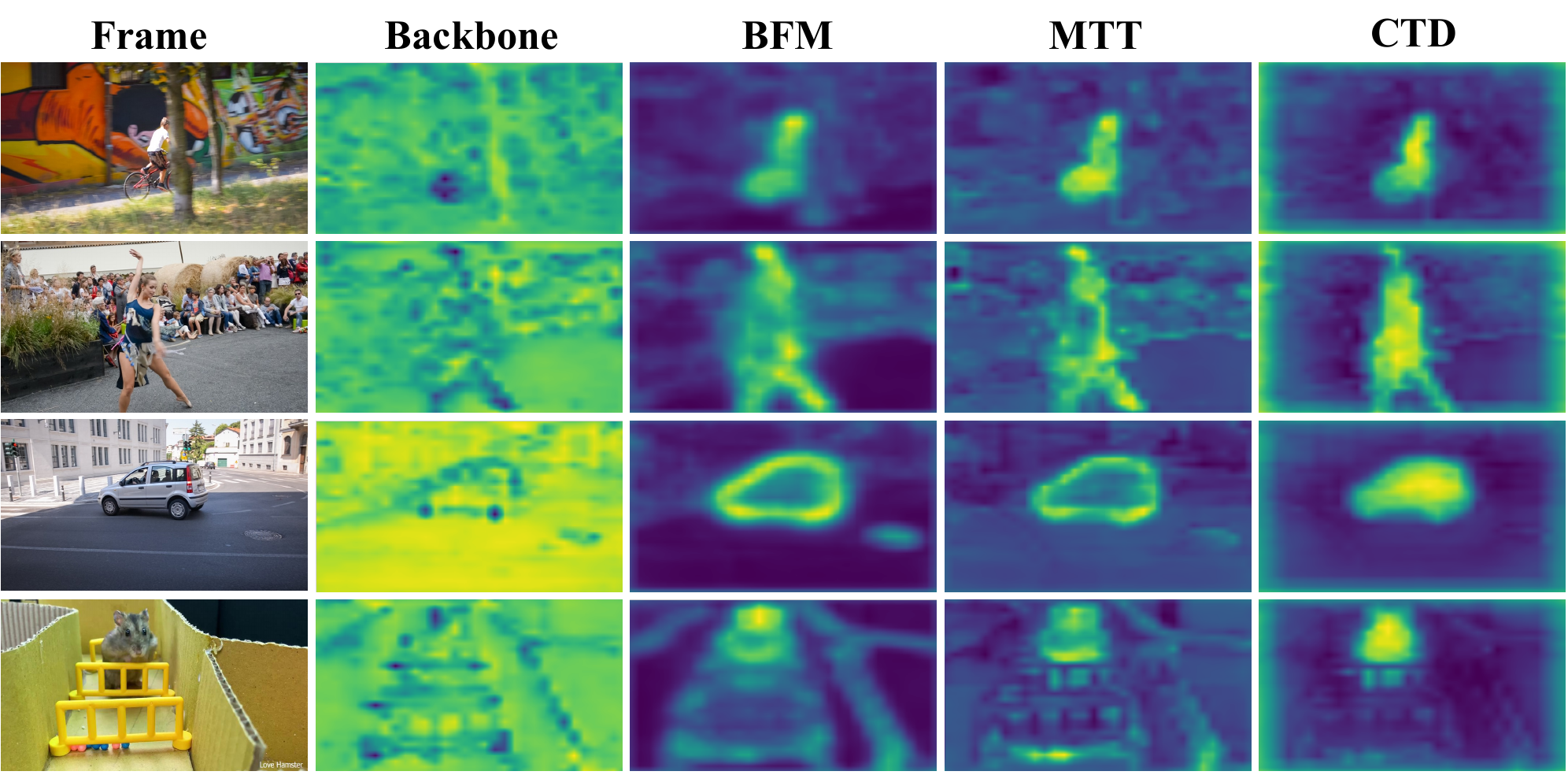} 
\end{tabular}
\caption{Visualizations of various components in our proposed MTNet. The column from left to right includes: the Input frames, Backbone visualization results, Bi-modal Fusion Module visualization results, Mixed Temporal Transformer visualization results, and Cascaded Transformer Decoder visualization results.}
\label{figure:visualization}
\vspace{-1mm}
\end{figure}

\subsection{Feature Visualization}
To enhance the understanding of our design components, we present visual comparisons for each element in Fig.~\ref{figure:visualization}. Initially, the features extracted from the backbone provide a broad representation of the video scene, encompassing both foreground objects and background elements. However, this inclusive representation may inadvertently introduce background noise and distractors, which may compromise the accuracy of object localization and segmentation in subsequent stages. To address these challenges, we have developed the Bi-modal Fusion Module (BFM), the Mixed Temporal Transformer (MTT), and the Cascaded Transformer Decoder (CTD). The BFM effectively merges appearance and motion information for each frame, facilitating the initial identification of foreground objects. Subsequently, the MTT refines this identification by enhancing foreground regions and attenuating background influences through inter-frame interactions. Finally, the CTD synthesizes multi-level features from both spatial and channel dimensions, yielding a feature map that is precise in object localization, maintains integrity in the target region, and delineates clear boundaries.


\begin{table}[ht]
\centering
\caption{Parameter analysis of our method and comparisons with other methods.}
\resizebox{65mm}{!}{\begin{tabular}{lccc}
\toprule[0.8pt]
Model & Model Size & Params & FLOPs \\
\midrule[0.6pt]
MATNet & 545.3M & 142.7M & 351.2G \\
FSNet & 390.9M & 102.3M & 149.4G \\
AMCNet & 600.3M & 157.1M & 292.2G \\
HFAN & 107.9M & 28.3M & 116.6G \\
Ours & 133.8M & 35.1M & 134.2G \\
\bottomrule[0.8pt]
\end{tabular}}
\vspace{-2mm}
\label{tab:parameters}
\end{table}

\subsection{Analysis of Computational Costs}
We present a comprehensive comparison of the computational costs of our MTNet with other leading methods in terms of model size, parameters and floating-point operations per second (FLOPs). As delineated in Tab.~\ref{tab:parameters}, MTNet exhibits an efficient balance between model complexity and computational expense. Despite having a moderate increase in the number of parameters compared to HFAN, MTNet achieves a significantly lower FLOP count than MATNet and AMCNet, indicating a more computationally economical architecture. Notably, this computational frugality does not compromise its performance or inference speed; MTNet delivers superior segmentation accuracy and frames per second (FPS), as evidenced in Table~\ref{tab:davis}. The optimal balance between  parameter count and FLOPs underscores the practicality of MTNet for deployment in real-world applications where computational resources are limited.

\subsection{Failure Cases Analysis}
\begin{figure}[t]
\vspace{-2mm}
\centering
\begin{tabular}{@{}c}
\includegraphics[width=0.92\linewidth]{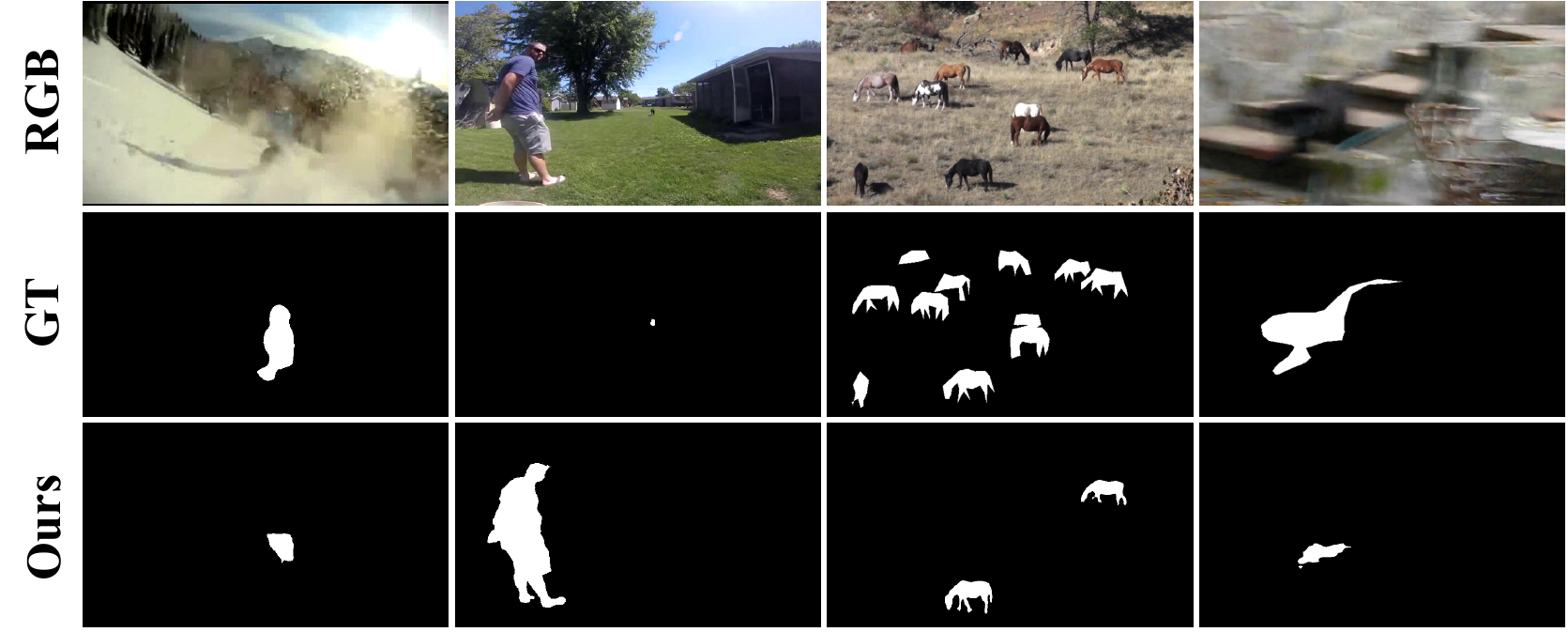} \\
\end{tabular}
\caption{Illustrative failures of MTNet in Unsupervised Video Object Segmentation: challenges in detecting blurred objects (1st and 4th columns), distinguishing foreground from prominent background elements (2nd column), and identifying salient objects in crowded scenes (3rd column).}
\label{figure:failure_case}
\vspace{-2mm}
\end{figure}

In Fig.~\ref{figure:failure_case}, we present several failure cases for analysis. As evident from the visualization, the model struggles in specific scenarios: when the primary object is obstructed (1st column) or when it becomes blurred (4th column). The model is unable to accurately locate, track, or segment the entire object under these conditions. In the 2nd column, a man positioned relatively close to the viewer appears on the left side of the image. Here, our model does not identify him as the primary object. In the 3rd column, which displays eleven houses entirely depicted as foregroundour approach successfully recognizes and segments only two, demonstrating a significant deviation from the ground truth.

This shortfall underlines the model's limitations in complex multi-instance circumstances, a common issue in instance-level datasets like DAVIS-19~\cite{caelles2019davis} and DAVSOD~\cite{fan2019shifting}. As articulated in~\cite{fan2019shifting}, the model's suboptimal handling of crowded scenes points to the necessity for improved discriminative feature learning, which is essential for distinguishing between salient and non-salient objects more effectively.


\section{Conclusion}

In the domain of computer vision, tackling unsupervised video object segmentation presents multifaceted challenges due to its inherent complexities. In addressing these challenges, this paper introduces MTNet, an innovative approach specifically designed to solve the challenges of unsupervised video object segmentation by adeptly synthesizing both motion and temporal cues.

The architecture of MTNet is grounded in three cornerstone components. Foremost, the Bi-modal Fusion Module plays a pivotal role, being specifically designed to deftly integrate appearance and motion features, thereby offering a more holistic representation. Subsequent to this, the Mixed Temporal Module is intricately designed to delve deep into the video, capturing the subtle temporal dynamics that are critical for effective segmentation. Concluding this trio is the Cascaded Transformer Decoder, a sophisticated entity that methodically refines multi-level features, culminating in the prediction of remarkably precise masks. Our extensive experimental evaluations demonstrate the effectiveness of MTNet, as it consistently outperforms state-of-the-art methods on benchmark datasets. 

While MTNet has proven to be effective for video segmentation tasks in both UVOS and VSOD across diverse datasets, it may face challenges in scenes with multiple similar objects, or where objects suddenly disappear or are heavily occluded. Additionally, although our model achieves a favorable balance between performance and speed on GPU devices, it is not yet optimized for deployment on mobile devices. Future work will focus on developing lighter model architectures and mobile-friendly operators. Addressing these specific challenges will be crucial for further improvements.

\printbibliography

\end{document}